\documentclass[a4paper,UKenglish,cleveref, autoref, thm-restate]{lipics-v2021}

\pdfoutput=1 %
\hideLIPIcs  %

\nolinenumbers
\usepackage{reledmac}
\usepackage{xspace}
\usepackage{color}
\usepackage{array}
\usepackage{underscore}
\usepackage{tipa}
\usepackage{cite}
\usepackage{stmaryrd}
\usepackage{mathabx}
\usepackage{tablefootnote}
\usepackage[binary-units=true,detect-weight=true]{siunitx}
\usepackage{rotating}

\providecommand{\qty}{\SI}

\usepackage[mathletters]{ucs}
\usepackage{rotating}

\usepackage{amsmath}
\usepackage{amssymb}
\usepackage{booktabs}
\usepackage{graphicx}
\usepackage{listings}
\usepackage{paralist}
\usepackage{hyperref}
\usepackage{breakurl}
\usepackage{footnote}

\newcolumntype{*}{>{\global\let\currentrowstyle\relax}}
\newcolumntype{^}{>{\currentrowstyle}}

\def\systemname#1{\textsf{#1}\xspace}

\def\th#1{\texttt{#1}\xspace}

\newcommand{\MML}{\systemname{MML}}

\newcommand*{\sym}{\mathcal}

\newcommand*{\SYMpremselLgb}{\ensuremath{\mathcal{L}}}
\newcommand*{\SYMpremselGnn}{\ensuremath{\mathcal{G}}}
\newcommand*{\SYMpremselKnn}{\ensuremath{\mathcal{K}}}
\newcommand*{\SYMpremselRnn}{\ensuremath{\mathcal{R}}}
\newcommand*{\SYMpremselNb}{\ensuremath{\mathcal{N}}}
\newcommand*{\SYMpremselMix}{\ensuremath{\mathcal{E}}}

\newcommand*{\SYMpremselMin}{\ensuremath{\mathcal{M}}}
\newcommand*{\SYMpremselBushy}{\ensuremath{\mathcal{B}}}

\newcommand*{\premselLgb}[2][]{\ensuremath{\SYMpremselLgb^{#1}_{#2}}}
\newcommand*{\premselGnn}[2][]{\ensuremath{\SYMpremselGnn^{#1}_{#2}}}
\newcommand*{\premselKnn}[2]{\ensuremath{\SYMpremselKnn^{\textrm{#1}}_{\textrm{#2}}}}

\newcommand*{\premselNb}[1]{\ensuremath{\SYMpremselNb_{\textrm{#1}}}}
\newcommand*{\premselMix}[2]{\ensuremath{\SYMpremselMix^{#1}_{#2}}}

\newcommand{\Str}[1]{\ensuremath{\mathcal{S}_{#1}}}

\title{MizAR 60 for Mizar 50}

\author{Jan Jakub\r{u}v}{Czech Technical University in Prague, Prague, Czech Republic}{jakubuv@gmail.com}{https://orcid.org/0000-0002-8848-5537}{}
\author{Karel Chvalovsk\'y}{Czech Technical University in Prague, Prague, Czech Republic}{karel@chvalovsky.cz}{https://orcid.org/0000-0002-0541-3889}{}
\author{Zarathustra Goertzel}{Czech Technical University in Prague, Prague, Czech Republic}{}{https://orcid.org/0000-0002-8458-2786}{}
\author{Cezary Kaliszyk}{University of Innsbruck, Austria and INDRC, Prague, Czech Republic}{cezary.kaliszyk@uibk.ac.at}{https://orcid.org/0000-0002-8273-6059}{}
\author{Mirek Ol\v{s}\'ak}{Institut des Hautes Études Scientifiques, Paris, France}{}{https://orcid.org/0000-0002-9361-1921}{}
\author{Bartosz Piotrowski}{Czech Technical University in Prague, Prague, Czech Republic}{bartoszpiotrowski@post.pl}{https://orcid.org/0000-0002-1699-018X}{}
\author{Stephan Schulz}{DHBW Stuttgart, Stuttgart, Germany}{}{https://orcid.org/0000-0001-6262-8555}{}
\author{Martin Suda}{Czech Technical University in Prague, Prague, Czech Republic}{martin.suda@cvut.cz}{https://orcid.org/0000-0003-0989-5800}{}
\author{Josef Urban}{Czech Technical University in Prague, Prague, Czech Republic}{}{https://orcid.org/0000-0002-1384-1613}{}

\authorrunning{Jakub\r{u}v et al.} %
\authorrunning{Jakub\r{u}v, Chvalovsk\'y, Goertzel, Kaliszyk, Ol\v{s}\'ak, Piotrowski, Schulz, Suda, Urban}

\Copyright{the authors} %



\category{} %

\relatedversion{} %
\EventEditors{John Q. Open and Joan R. Access}
\EventNoEds{2}
\EventLongTitle{42nd Conference on Very Important Topics (CVIT 2016)}
\EventShortTitle{CVIT 2016}
\EventAcronym{CVIT}
\EventYear{2016}
\EventDate{December 24--27, 2016}
\EventLocation{Little Whinging, United Kingdom}
\EventLogo{}
\SeriesVolume{42}
\ArticleNo{23}
\begin{document}
\maketitle
\begin{abstract}
As a present to Mizar on its 50th anniversary, we develop an AI/TP system that
automatically proves about \qty{60}{\percent} of the Mizar theorems in the hammer setting. We also
automatically prove \qty{75}{\percent} of the Mizar theorems when the automated provers are
helped by using only the premises used in the human-written Mizar proofs.  We
describe the methods and large-scale experiments leading to these results. This
includes in particular the E and Vampire provers, their ENIGMA and Deepire
learning modifications, a number of
learning-based premise selection methods,
and the incremental loop that interleaves growing a corpus of
millions of ATP proofs with %
training increasingly strong AI/TP systems on them. We also present a selection of Mizar problems that were proved automatically. %
\end{abstract}

\section{Introduction: Mizar, MML, Hammers and AITP}
\label{introduction}

In recent years, methods that combine machine learning (ML),
artificial intelligence (AI) and automated theorem proving
(ATP)~\cite{DBLP:books/el/RobinsonV01} have been considerably
developed, primarily targeting large libraries of formal mathematics
developed by the ITP community. This ranges from \emph{premise
  selection} methods~\cite{abs-1108-3446} and \emph{hammer}~\cite{hammers4qed} systems to
developing and training learning-based \emph{internal guidance} of ATP
systems such as E~\cite{Schulz13,SCV:CADE-2019} and Vampire~\cite{Vampire} on the thousands to
millions of problems extracted from the ITP libraries. Such large ITP
corpora have further enabled research topics such as \emph{automated
  strategy invention}~\cite{blistr} and \emph{tactical guidance}~\cite{GauthierKUKN21},
learning-based \emph{conjecturing}~\cite{UrbanJ20}, \emph{autoformalization}~\cite{DBLP:conf/itp/KaliszykUV17,Wang18},
and development of metasystems that combine learning and reasoning in
various feedback loops~\cite{US+08}.

Starting with the March 2003 release\footnoteA{\url{http://mizar.uwb.edu.pl/forum/archive/0303/msg00004.html}} of the MPTP
system~\cite{Urb03} and the first ML/TP and
hammer experiments over it~\cite{Urb04-MPTP0}, the Mizar
Mathematical
Library~\cite{BancerekBGKMNP18,BancerekBGKMNPU15,mizar-in-a-nutshell}
(\MML) and its subsets have as of 2023 been used for twenty years for this research,
making it perhaps the oldest and most researched AI/TP resource in the last two decades.

\subsection{Contributions}

The last large \emph{Mizar40} evaluation~\cite{KaliszykU13b} of the AI/TP methods over
MML was done almost ten years ago, on the occasion of 40 years of
Mizar. Since then, a number of strong methods have been developed in
areas such as premise selection and internal guidance of ATPs. In this
work, we therefore evaluate these methods in a way that can be
compared to the Mizar40 evaluation, providing an overall picture
of how far the field has moved. Our main results are: %
\begin{enumerate}
\item Over \qty{75}{\percent} of
  the Mizar toplevel lemmas can today be proved by AI/TP systems
 when the premises for the proof can be selected from the library either by a human or a machine.
 This should be compared to \qty{56}{\percent} in Mizar40 achieved on the same version of the MML.
 Over 200 examples of the automatically obtained proofs are analyzed on our web
 page.\footnoteA{\url{https://github.com/ai4reason/ATP_Proofs}}
\item \qty{58.4}{\percent} of the Mizar toplevel lemmas can be proved today
  without any help from the users, i.e., in the large-theory
  (hammering) mode. This should be compared to about \qty{40.6}{\percent} achieved on
  the same version of the MML in Mizar40. In both cases, this is done by a
  large portfolio of AI/TP methods which is limited to \qty{420}{\second} of CPU time.
\item Our strongest single AI/TP method alone now proves in \qty{30}{\second} \qty{40}{\percent}
  of the %
  lemmas in the hammering mode, i.e., reaching the
  same strength as the full \qty{420}{\second} portfolio in Mizar40.
\item Our strongest \emph{single} AI/TP method now proves in \qty{120}{\second} \qty{60}{\percent}
  of the toplevel lemmas in the human-premises (\emph{bushy}) mode (Section~\ref{s:mml1382}), i.e., outperforming the union of \emph{all} methods developed in Mizar40 (\qty{56}{\percent}).
\item We show that our strongest method transfers to a significantly newer
  version of the MML which contains a lot of new terminology and
  lemmas. In particular, on the new \num{13370} theorems coming from the new
  242 articles in MML version \num{1382}, our strongest method outperforms
  standard E prover by \qty{58.2}{\percent}, while this is only \qty{56.1}{\percent} on the
  Mizar40 version of the library where we do the training and
  experiments. This is thanks to our development and use of
  \emph{anonymous}~\cite{JakubuvCOP0U20} logic-aware ML methods that learn only from
  the structure of mathematical problems.  This is unusual in
  today's machine learning which is dominated by large language models that
  typically struggle on new
  terminology.
\end{enumerate}

\subsection{Overview of the Methods and Experiments}
\label{sec:overview}
The central methods in this evaluation are internal guidance provided by
the ENIGMA (and later also Deepire) system, and premise
selection methods. We have also used several additional approaches such as
many previously invented strategies and new methods for constructing
their portfolios, efficient methods for large-scale training on
millions of ATP proofs, methods that interleave multiple runs of ATPs
with restarts on ML-based selection of the best inferred clauses
(\emph{leapfrogging}), and methods for minimizing the premises needed for the
problems by decomposition into many ATP subproblems. These methods are
described in Sections~\ref{enigma}, \ref{premise}, and \ref{sec:stratsports}, after introducing the MML in Section~\ref{corpus}.
Section~\ref{Results} describes the large-scale evaluation and its final results,
and Section~\ref{Proofs} showcases the obtained proofs.

\section{The Mizar Mathematical Library and the Mizar40 Corpus}
\label{corpus}

Proof assistant systems are usually developed together with their respective proof libraries. This allows evaluating
and showcasing the available functionality. In the case of Mizar \cite{BancerekBGKMNPU15},
the developers have very early decided to focus on its library, the MML (Mizar Mathematical
Library) \cite{BancerekBGKMNP18}. This was done by establishing a dedicated library committee
responsible for the evaluation of potential Mizar articles to be included, as well as for maintaining
the library. As a result, the MML became one of the largest libraries of formalized mathematics today.
It includes many results absent from those derived in other systems, such as
lattices~\cite{BancerekR02} and random-access Turing machines~\cite{SCM}.

All the data gathered and evaluations performed in the paper (with the
exception of version-transfer in Section \ref{s:mml1382}) use the same
Mizar library version as the previous large evaluation
\cite{KaliszykU13b} and all subsequent evaluation papers. This allows
us to rigorously compare the methods and evaluate the
improvement. That version of the library, MML 1147, when exported to
first-order logic using the MPTP export \cite{Urban06} corresponds to
\num{57897} theorems including the unnamed toplevel lemmas. For a rigorous evaluation in the hammering scenario, we will
further split this dataset into several training and testing parts in
Section~\ref{sec:trainprems}.
\section{ENIGMA: ATP Guidance and Related Technologies}
\label{enigma}

ENIGMA~\cite{JakubuvU17,JakubuvU18,ChvalovskyJ0U19,10.1007/978-3-030-29026-9_21,JakubuvU19,JakubuvCOP0U20,GoertzelCJOU21,GoertzelJKOPU22,Goertzel20}
stands for ``\emph{\underline{E}fficient Lear\underline{n}ing Based
  \underline{I}nference \underline{G}uiding \underline{Ma}chine}''. It is the first learning-guided ATP that
in 2019 achieved large improvements over state-of-the-art saturation ATPs~\cite{JakubuvU19}, and the main ingredient of the work reported here.
This section summarizes previously published research on ENIGMA and also the related
methods that were used to undertake the large-scale experiments done %
here (Section~\ref{Results}).

\subsection{Saturation Theorem Proving Meets Machine Learning}
\label{sec:atp-ml}

\textbf{Saturation Provers:} State-of-the-art automated theorem provers, like E Prover~\cite{Sch02-AICOMM}
and Vampire~\cite{Vampire}, perform the search for a contradiction, first
translating the input first-order logic problem into a refutationally
equivalent set of clauses.
Then the prover operates the proof search using the \emph{given clause algorithm}.
In this algorithm, the proof state is split into two subsets, the set $P$ of
\emph{processed clauses}, and the set $U$ of unprocessed clauses. Clauses in
$U$ are ordered by a heuristic evaluation function. In each iteration of the
main loop, the (heuristically) best clause in $U$ is picked. This \emph{given
clause} $g$ is then simplified with respect to all clauses in $P$. If it is not
redundant, it is used in turn to simplify all clauses in $P$. After that, all
generating inferences between $g$ and the remaining clauses in $P$ are
performed. Both the newly generated clauses and the simplified clauses from $P$
are then completely simplified with respect to $P$, heuristically evaluated,
and added to $U$. This process continues until the empty clause emerges (or
until the system runs out of resources).

\textbf{Training Data:} As of E~1.8~\cite{Schulz:LPAR-2013}, E maintains an internal proof
object~\cite{SS:APPA-2015} which allows it to
inspect all proof clauses and %
designate all clauses that have been selected
for processing and are part of the proof, as \emph{positive training
examples}. All clauses that have been selected for processing, but not
contributed to the proof, are designated as \emph{negative training
examples}. Clauses that have not been processed at all are neither
positive nor negative, reducing the total number of training examples
to typically thousands of processed clauses, as opposed to millions of
clauses generated.
E allows the user to request the actual proof object, or to provide
any combination of positive and negative training examples. Examples
are provided in separate batches and are also annotated as positive
or negative for easy processing.

\textbf{ML-Based Selection:} Selection of the right given clause is critical in E, and an
obvious point for the use of machine learning (ML). The positive and negative examples
are extracted from previous successful proof searches,
and a machine learning model is trained to %
score the generated clauses or to classify them as useful ($\boxplus$) or useless
($\boxminus$). %
E Prover selects the given clause from a priority queue, where the unprocessed
clauses are sorted by various heuristics.
ENIGMA extends E Prover with an additional queue where clauses positively
classified by the ML model are prioritized.
The ENIGMA queue is used together with the standard E selection mechanisms,
typically in a cooperative way where roughly half of the clauses are selected by
ENIGMA.
This approach proved to be the most efficient in practice.

\begin{figure}[t]
\begin{center}
\begin{minipage}[]{0.52\textwidth}
  \includegraphics[width=1\textwidth]{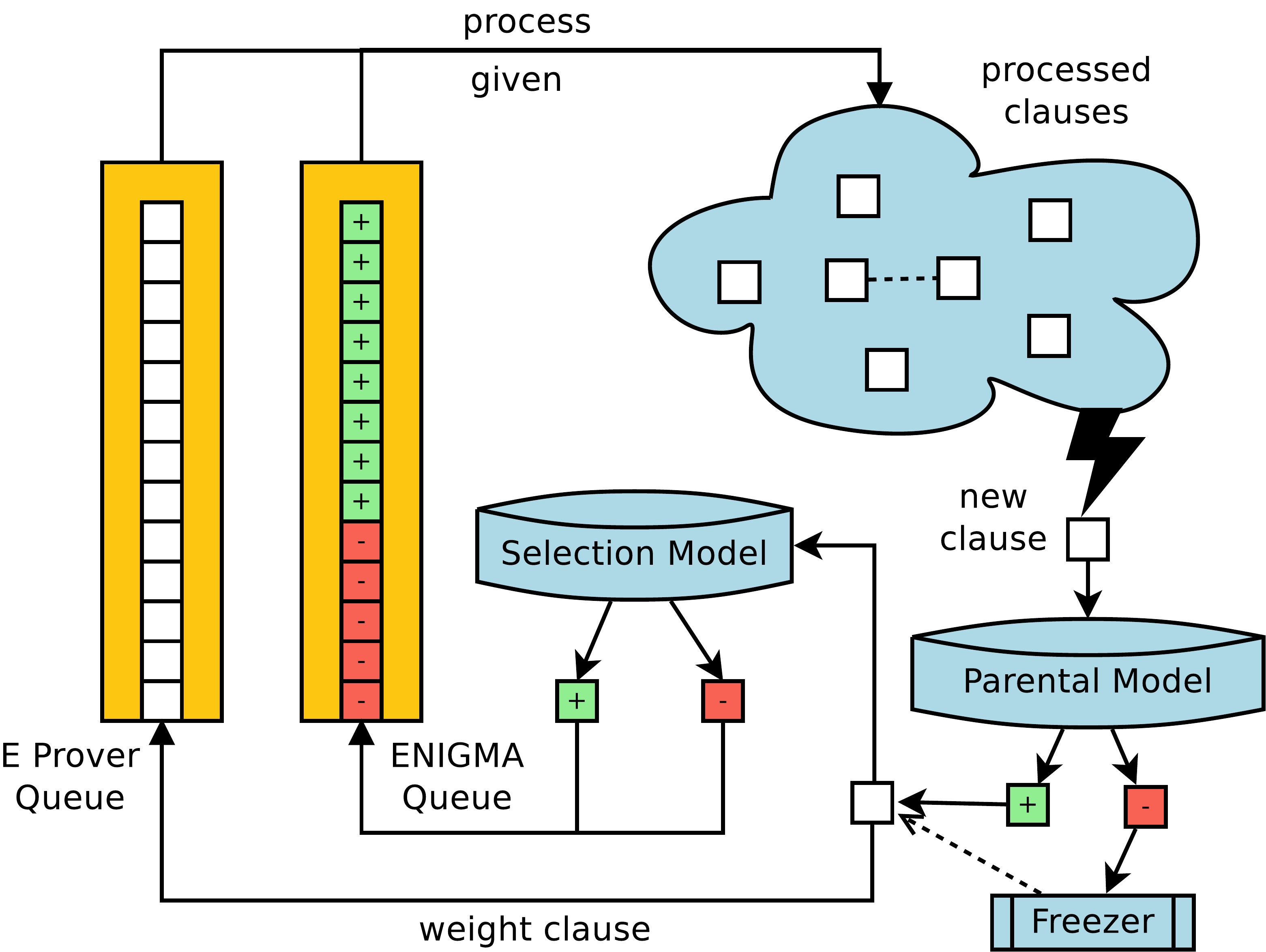}
\end{minipage}
\hfill
\begin{minipage}[]{0.16\textwidth}
  \includegraphics[width=1\textwidth]{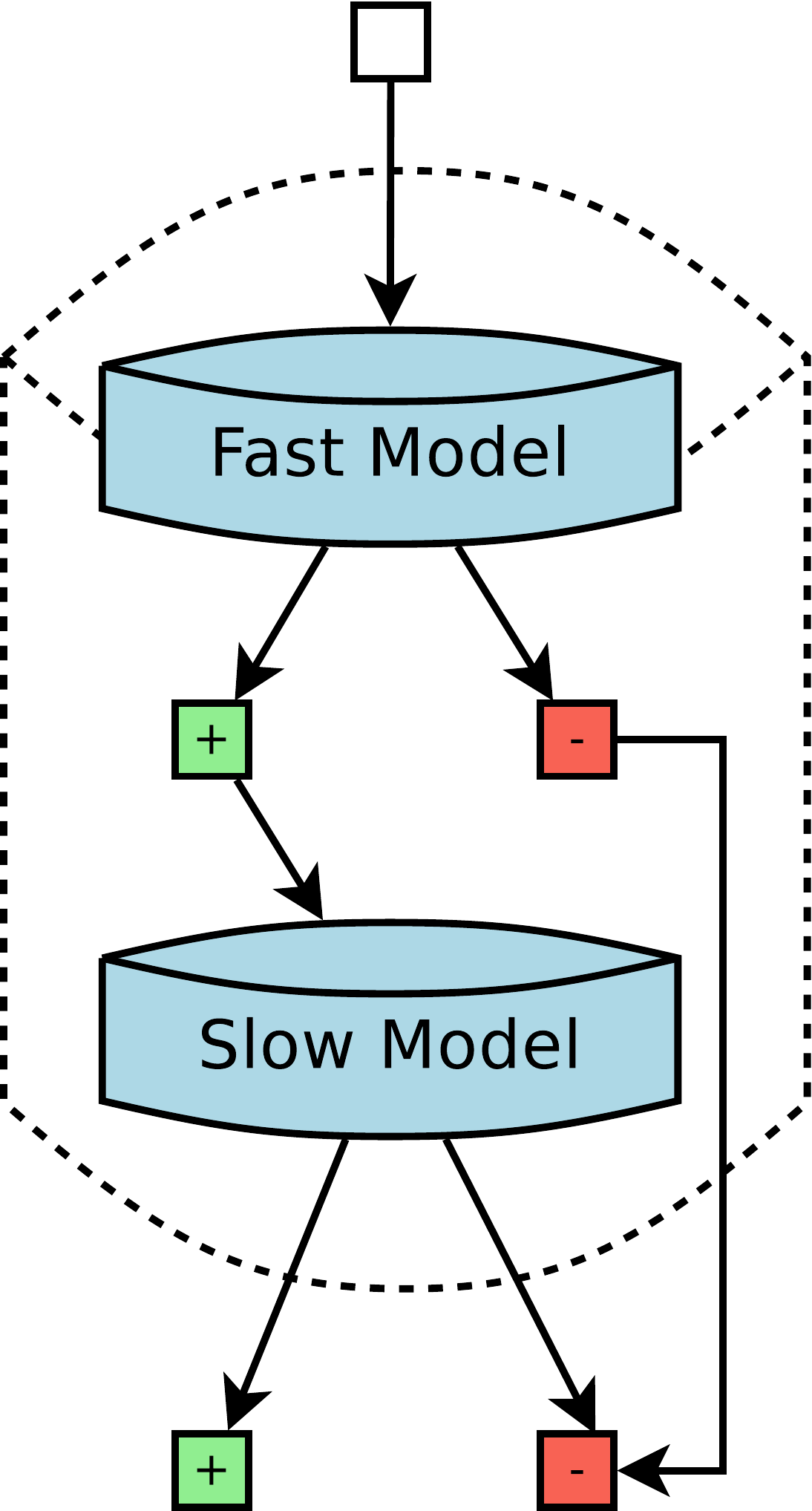}
\end{minipage}
\hfill
\begin{minipage}[]{0.24\textwidth}
\includegraphics[width=1\textwidth]{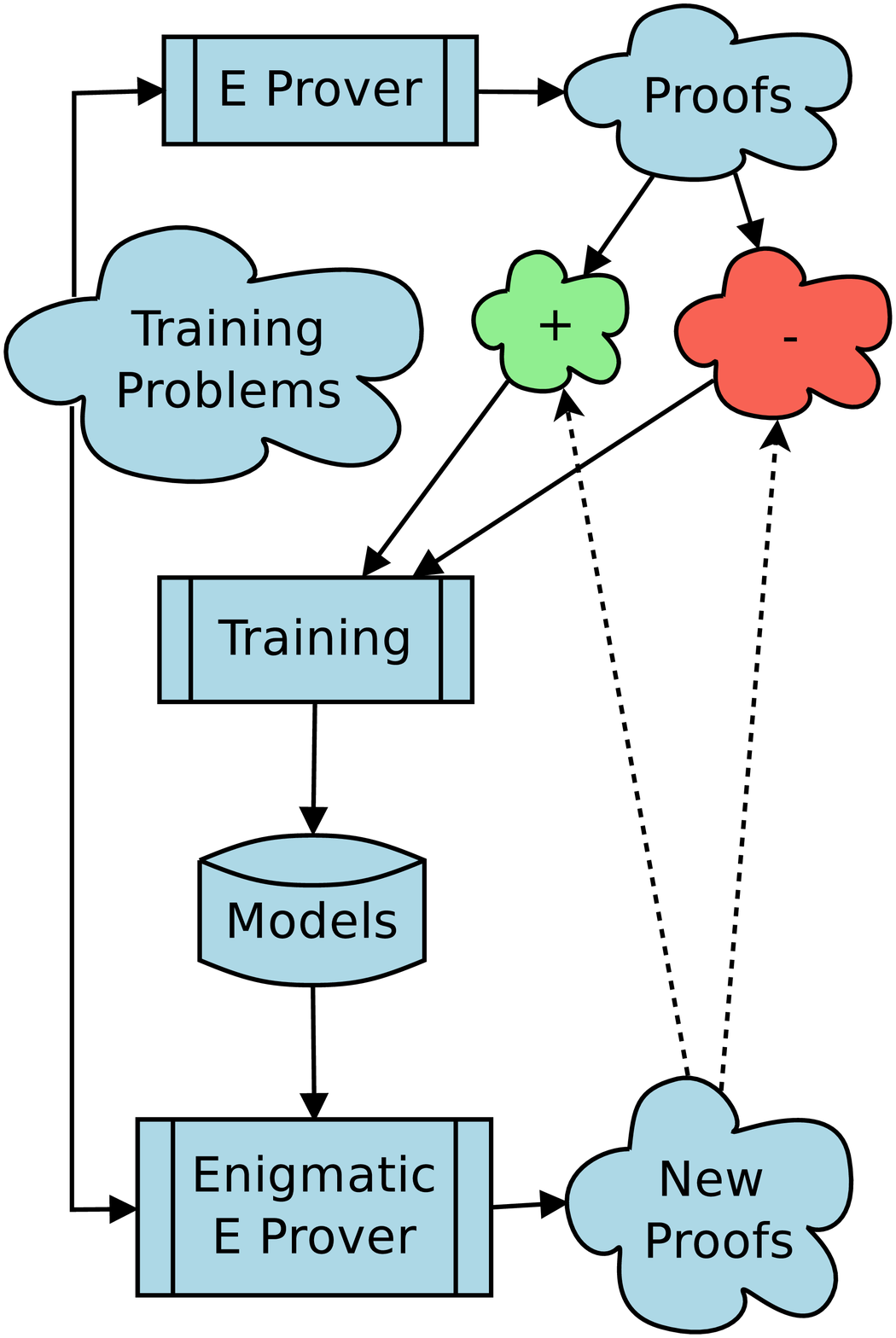}
\end{minipage}
\end{center}
\caption{Schema of E Prover with ENIGMA (left), of a two-phase selection model
(middle), and of the prove-learn feedback loop (right).}
\label{fig:enigmatic-eprover}
\end{figure}

\textbf{Parental Guidance:} Later ENIGMA~\cite{GoertzelCJOU21} introduced
learning-based \emph{parental guidance}, which addresses the quadratic
factor when doing \emph{all} possible inferences among the processed
clauses in classical saturation-based provers. Instead, an ML model is trained to prevent inferences
between the parent clauses that are unlikely to meaningfully interact.
When such an inference is %
recognized by the model as useless with a
high degree of confidence, the child clause is not inserted into the set of
unprocessed clauses $U$ but its processing is postponed.
To maintain completeness, the clause can not be directly discarded
since the ML model might be mistaken.
Instead, the clause is put into a ``freezer'' from which it can be retrieved in
the case the prover runs out of unprocessed clauses.
As opposed to the above clause selection models, this method affects the
standard E selection mechanism because the clause is not inserted into any
queue.
ENIGMA clause selection models and parental (\emph{generation}) models can be successfully
combined.
This is schematically illustrated in Figure~\ref{fig:enigmatic-eprover} (left).

\textbf{Multi-Phase ENIGMA:} ML-based multi-phase clause selection was introduced in~\cite{GoertzelCJOU21}
to deal with computationally expensive (\emph{slow}) ML models, like graph neural networks (GNNs).
In a \emph{two-phase} selection model, a \emph{faster} model is used for
preliminary clause filtering, and only the clauses that pass %
are evaluated by the
slower model.
The fast model is expected to over-approximate on positive classes so that
only clauses classified with high confidence as negatives are rejected. When parental guidance is added to the mix,
this leads to a \emph{three-phase} ENIGMA.
This is schematically illustrated in Figure~\ref{fig:enigmatic-eprover} (middle).
\emph{Aggressive forward subsumption} is an additional logic-complete pruning method
based on efficient subsumption indexing in E~\cite{DBLP:conf/birthday/Schulz13}.
We use it to eliminate many redundant generated clauses
before calling more expensive ML methods (GNN) for clause evaluation. For the effect of such methods, see some of the 3-phase ENIGMA examples in Section~\ref{Proofs}.

\textbf{Training:} Strong ENIGMAs are typically developed in many prove-learn feedback loops~\cite{US+08} that proceed as follows.
(1) The training data $\sym{T}$ are curated from (previous) successful proof searches.
(2) A model $\sym{M}$ is trained on data
  $\sym{T}$ to distinguish positive from negative clauses.
(3) The model $\sym{M}$ is run with the ATP (E), usually in \emph{cooperation}
with the strategy used to obtain the training data.
Then we go to step (1) with the new data obtained in step (3).
The loop, illustrated in Figure~\ref{fig:enigmatic-eprover} (right),
can be repeated as long as new problems are proved. We run this loop for several months in this work.

\subsection{Gradient Boosted Decision Tree Classifiers and Features}
\label{sec:gbdt}

ENIGMA supports classifiers based on Gradient Boosted Decision Trees (GBDTs).
In particular, we experiment with XGBoost~\cite{Chen:2016:XST:2939672.2939785}
and LightGBM~\cite{LightGBM}. %
Both frameworks are
efficient and can handle large data well both in training and evaluation.
For learning, we represent first-order clauses by numeric \emph{feature vectors}.
A decision tree is a binary tree with nodes labeled by conditions on the values of the feature vectors.
Given a clause, the tree is navigated to the leaf where the clause evaluation is stored.
Both frameworks work with a sequence (\emph{ensemble}) of several trees, constructed in a
progressive way (\emph{boosting}).
The frameworks differ in the underlying algorithm for the construction of
decision trees.
XGBoost constructs trees level-wise, while LightGBM leaf-wise.
This implies that XGBoost trees are well-balanced.
On the other hand, LightGBM can produce much deeper trees, and the tree depth
limit is indeed an important learning meta-parameter that can be optimized.

ENIGMA extracts various syntactic information from a first-order clause and
stores them in the feature vector of the clause.
Given a finite set of features, each feature is assigned an index in the feature
vector, and the corresponding feature value is stored at this index.
For example, a typical clause feature is the clause length.
ENIGMA supports the following.
\textbf{Vertical Features}
are constructed by traversing the clause syntax tree and collecting
all top-down oriented symbol paths of length $3$.
Additionally, to abstract from variable names and to deal with possible
collisions of Skolem symbols, all variables are replaced by a special
name $\odot$ and all Skolem symbols by $\ocoasterisk$.
\textbf{Horizontal Features} introduce for every term
  $f(t_1,\dots,t_n)$, a new feature ${f(s_1,\dots,s_n)}$, where $s_i$ is the top-level symbol of $t_i$.
\textbf{Count Features} include the clause length, literal counts, and similar statistics.
\textbf{Conjecture Features} embed the conjecture to be proved in the feature
   vector.
   Thusly, ENIGMA is able to provide goal specific predictions.
\textbf{Parent Features} represent a clause by features (concatenated or summed) of its parents.
\textbf{Feature Hashing}
   is an important step towards large data in ENIGMA~\cite{ChvalovskyJ0U19}.
It significantly reduces the feature vector size and thusly allows handling of
larger data.
Each feature is represented by a unique string identifier. %
This string is passed through the hashing function %
and the hash modulo the selected \emph{hash base} is used as the feature
index.
\textbf{Symbol Anonymization} allows to abstract from specific symbol
   names~\cite{JakubuvCOP0U20}.
During the extraction of clause features, all symbol names are replaced by
symbol arities, keeping only the information whether the symbol
is a function or a predicate.
In this way, a decision tree classifier does not depend on symbol names, at
the price of
symbol collisions, which are however empirically mitigated by collecting longer paths as features.

\subsection{Graph Neural Network (GNN) Classifiers}
\label{enigma:gnn}

Anonymizing \emph{graph neural networks} provide an alternative approach for abstracting from specific terminology.
ENIGMA uses~\cite{JakubuvCOP0U20} a symbol-independent GNN
architecture initially developed for guiding tableaux search~\cite{OlsakKU20} implemented in TensorFlow~\cite{tensorflow2015-whitepaper}.
A set of
clauses is directly represented by a hypergraph with three
kinds of nodes for clauses, subterms/literals, and symbols.
Relationships among the objects are represented by various graph edges, which
allow the network to distinguish different symbols while abstracting from their
names.

The GNN layers perform
message passing across the edges, so the information at every node can get to its neighbors.
This allows the network to see how
the symbols are used without knowing their names. We always classify
the new clauses together with the initial clauses which provide
the context %
for the meaning of the anonymized symbols.
During the ATP evaluation, predictions of hundreds of generated clauses are computed at once in larger batches, with the context given both by the initial and the processed clauses.
The context can be either \emph{fixed}, containing an initial segment of the initial and processed clauses, or
it can be a \emph{shifting context} using a window of clauses with the best GNN
evaluation. %

\subsection{Additional Related Techniques}\label{s:related}

\textbf{GPU Server Mode} allows using GPUs for real-time
   evaluation~\cite{GoertzelCJOU21}.
To reduce the GPU overhead of model loading, we developed a Python GPU server,
with preloaded models that can distribute the evaluation over several GPUs.
E Prover clients communicate with the server via a network socket.
We
fully utilize our physical server%
\footnoteA{36 hyperthreading Intel(R) Xeon(R) Gold 6140 CPU @ \qty{2.30}{\giga\hertz}
cores, \qty{755}{\giga\byte} of memory, and 4 NVIDIA GeForce GTX 1080 Ti GPUs.}
when we run 160 instances of E prover in parallel.
Running both the server and clients on the same machine reduces the network
communication overhead.

\textbf{Leapfrogging} addresses the problem of evolving context when new given
   clauses are selected~\cite{DBLP:conf/tableaux/ChvalovskyJOU21}.
We run ENIGMA with a given abstract limit and generate a larger set of clauses. Then we run
a premise selection on these generated clauses (e.g., only
processed clauses), take the good clauses, and use them as input for a
new ENIGMA run.
A related \emph{split/merge} method involves repeatedly
splitting the generated clauses  into
components that are run separately and then merged with premise selection. This is inspired by the idea that harder problems consist of components
that benefit from such divide-and-conquer approaches.

\textbf{Deepire}
is an extension~\cite{DBLP:conf/cade/000121a,DBLP:conf/frocos/Suda21}
of Vampire \cite{Vampire} by machine-learned clause selection guidance,
generally following the ENIGMA-style methodology. It is distinguished
by its use of recursive neural networks for classifying the generated clauses based solely on their derivation history.
Thus Deepire does not attempt to read ``what a clause says'', but only bases its decisions
on ``where a clause is coming from''. This allows the clause evaluation to
be particularly fast, while still being able to recognize and promote useful clauses,
especially in domains with distinguished axioms which reappear in many problems.
\section{Learning Premise Selection From the \MML}
\label{premise}

When an ATP is used
over a large ITP library, typically only a small fraction of the %
facts are relevant for proving a new conjecture.
Since giving too many redundant
premises to the ATP significantly decreases the chances of proving the
conjecture, premise selection is a critical task.
The most efficient premise selection methods use \textit{data-driven} or
\textit{machine-learning} approaches. If $T$ is a set of theorems with their
proofs and $C$ is a set of conjectures without proofs, the task is to learn a
(statistical) model from $T$, which for each conjecture $c \in C$ will rank (or
select a subset of) its available premises according to their relevance for
producing an ATP proof of $c$. Two %
main machine learning settings can be
used. %
In \textbf{Multilabel classification}, premises used in the
		proofs are treated as opaque labels and a machine learning model is trained to label conjectures based on their features.
    \textbf{Binary classification} aims to recognize
    pairwise-relevance of the (\textit{conjecture,
        premise}) pairs, i.e.\ to estimate the chance of a
		premise being relevant for proving the conjecture based on the
		features of both the conjecture and the premise.

The first setting is suitable for simpler, fast ML methods, like
$k$-NN or Naive Bayes -- these are described in Section \ref{sec:knn-nb}. The
second setting (Section~\ref{sec:binary}) allows using more powerful ML architectures, like GBDTs and GNNs  %
(Sections~\ref{enigma:gnn} and \ref{sec:gbdt}). However, this setting
also requires selecting \textit{negative examples} for training~\cite{PiotrowskiU18}, which increases its complexity.

\subsection{Multilabel Premise Selection (\SYMpremselKnn, \SYMpremselNb, \SYMpremselRnn)}
\label{sec:knn-nb}

Naive Bayes and $k$-nearest neighbors
were the strongest selection
methods in the %
Mizar40 evaluation~\cite{KaliszykU13b}.
In this work, we improve them and apply them together with newer methods.

\textbf{\textit{k}-NN (\SYMpremselKnn):} The $k$-nearest neighbours algorithm, when applied to premise selection,
chooses $k$ facts closest to the conjecture in the feature space and selects their dependencies.
Already known modifications of the standard $k$-NN %
include considering the number of dependencies of facts (proofs with more dependencies
are longer and thus less important) as well as TF-IDF (rare features are more important) \cite{EasyChair:74}.
Additionally, we realize that we do not need to fix the $k$. Instead,
we consider a small $k$ and if the number of scored dependencies %
is too low,
we increase the $k$ and update the dependencies. %
This is
repeated until the requested number of predictions is obtained. The
$k$-NN-based predictions with fixed $k$ will be denoted, e.g., by \premselKnn{}{512}, while
with variable $k$ this will be \premselKnn{fea}{var}, where \textit{fea} specifies the features used.

\textbf{Naive Bayes (\SYMpremselNb):} The sparse Naive Bayes algorithm estimates the relevance of a fact $F$
by the conditional probability of $F$ being useful (estimated from past proof statistics)
under the condition of the features being present in the conjecture (again estimated from
statistics). We %
also consider \emph{extended
features} of $F$, i.e.,  features of $F$ and features of facts proved using $F$.
Together with premise selection-specific weights this improves on the basic Naive Bayes
and has already been used in HolyHammer and later Sledgehammer. A complete derivation
of the algorithm is discussed in %
\cite{BlanchetteGKKU16}.
The Naive Bayes predictions will be denoted by \premselNb{fea}.

These %
algorithms can %
be parametrized by
more complex %
\emph{features}. We considered:
\texttt{cp} for constants and paths (Section~\ref{sec:gbdt}) in the term graph,
\texttt{sub} for subterms,
\texttt{au} for anti-unification %
features \cite{DBLP:conf/ijcai/KaliszykUV15},
\texttt{eni} for online ENIGMA features discussed in Section~\ref{sec:gbdt}
and \texttt{uni} for the union of all above.
Finally, these algorithms also support the \emph{chronological} mode, which in the learning
phase discards proofs that use facts introduced after the current conjecture in the Mizar
canonical order (\texttt{MML.LAR}). This slightly weakens the algorithms, but is compatible
with the previous Mizar40 premise selection evaluation \cite{KaliszykU13b}. These will be marked by $^{chrono}$.

\textbf{Dependent Selection with RNNs (\SYMpremselRnn):}
Premise selection methods were originally mainly based on %
\emph{ranking} the facts  \emph{independently}
with respect to the conjecture. %
The highest ranked facts
are then used as axioms and given to the ATP systems together with the
conjecture. Such approaches (used also with GBDTs), although useful and successful, do not take into
account
that the premises are
\emph{not} independent of each other. %
Some premises complement each other better when proving a
particular conjecture, while some highly-ranked premises might be just minor
variants of one another. %
Recurrent neural network (RNN) encoder-decoder
models~\cite{Cho14}
and transformers~\cite{DBLP:journals/corr/VaswaniSPUJGKP17} (language models) turn out to be suitable ML %
architectures for modeling such implicit dependencies. %
Such models have been traditionally developed for natural language processing,
however, recently they are also increasingly used in symbolic
reasoning tasks \cite{Evans18,Wang18,abs-1911-04873,Lample20,abs-2301-11479}, including premise selection
\cite{PiotrowskiU20}.

\subsection{Premise Selection as Binary Classification (\SYMpremselLgb, \SYMpremselGnn)}
\label{sec:binary}

\textbf{Gradient Boosted Decision Trees (\SYMpremselLgb{}):} We use GBDTs (LightGBM) also for premise selection in the binary mode.
They are faster to train than the deep
learning methods, perform well with unbalanced training sets, and handle well sparse features.
We fix the LightGBM hyperparameters here based on %
our previous experiments with  applying GBDTs to premise selection~\cite{PiotrowskiU18}.
In the binary setting, the GBDT scores the pairwise
relevance of the conjecture and a candidate premise. Because the number of
possible candidates is large (all preceding facts in the large ITP library), we first use the cheaper $k$-nearest neighbors
algorithm to pre-filter the available premises.
The predictions from LightGBM will be denoted as \premselLgb{} below.

\textbf{Dependent Selection with GNNs (\SYMpremselGnn{}):}
The message-passing GNN architecture
described in Section~\ref{enigma:gnn}
can also be applied to premise selection.
Like RNNs, it can also take into account the dependencies between premises.
As the GNN is relatively slow, we
will use it in combination with a simpler premise selection method, such as $k$-NN, preselecting 512 facts. %
We will denote GNN predictions by \premselGnn{} below.
Both \premselLgb{} and \premselGnn{}, can be indexed with the threshold on the
score (like $\premselLgb{0.1}$ or $\premselGnn{-1}$), used to
differentiate useful and useless clauses.

\subsection{Ensemble Methods for Premise Selection (\SYMpremselMix)}

There are several ways how we can combine the premise selection methods discussed in previous subsections.
Naturally, using different methods for different strategies works well, however, we also found that
combining the predictions obtained from several %
methods and using them
for a single prover run gives good and complementary results.
Since prediction scores resulting from different algorithms are often incomparable \cite{OpitzM99}, we only
use the rankings produced by the various methods and based on this we create a combined ranking. We have
compared several ways to combine rankings in previous work \cite{EasyChair:74} and found that several averages
work well: arithmetic mean, minimum, and geometric mean, with the harmonic mean giving experimentally
the best results. Additionally, we add weights to the different combined methods. The weights give more priority to
a stronger prediction method, but allow it to benefit from the simpler ones overall (by picking up some lost facts).
Given %
predictions from $n$ different methods and method weights $w_1,\ldots,w_n$, assume
that a fact has been ranked as $r_1$-th by the first method until and $r_n$-th by the last one. Then, the
ensemble method would give that fact a score of
$1/\sum_{i=1}^{n}\frac{w_i}{r_i}$.
The scores
of the facts obtained in this way are sorted, to get a ranking of all facts.
The ensemble predictions will be denoted by \SYMpremselMix, with methods and their weights in the super and subscript,
for example \premselMix{\SYMpremselKnn,\SYMpremselNb,\SYMpremselGnn}{0.25,0.25,0.5}.

\subsection{Subproblem Based Premise Minimization (\SYMpremselMin)}
\label{subproofs}

The proof dependencies obtained by successful ATP runs typically
perform better as data for premise selection than the dependencies
from the human-written ITP
proofs~\cite{holyhammer,hammers4qed}. However, some Mizar proofs are
hundreds of lines long and it is so far unrealistic to raise the \qty{75}{\percent}
ATP performance obtained here in the bushy setting to a number close
to \qty{100}{\percent}. This means that if we used only ATP-based premise data, we
would currently miss in the premise selection training \qty{25}{\percent} of the
proof dependency information available in the MML.

To remedy that, we
newly use here \emph{subproblem based premises}.
The idea behind this is that a theorem with a longer Mizar proof consists of a series of natural deduction steps that typically have to be justified.
Once ATP proofs of all such steps (we call them subproblems) for a
given toplevel theorem are available, they can be used to prune the
(overapproximated) set of human-written premises of the theorem. Such minimization also increases the chance of proving the theorem directly.
In more detail, we consider  the following approaches: %
  (1) Use the premises from only ATP-proved subproblems, ignoring unproved subproblems. %
(2) Add to (1) all explicit Mizar premises of the theorem (possibly ignoring some background facts).
(3) Add to (2) also the (semi-explicit) definitional expansions detected by the natural deduction module.
(4) Add to (3) also some of the background premises, typically those ranked high by the trained premise selectors.
When using (1) and (3), we were able to prove more than \num{1000} hard theorems (see Table~\ref{t:timeline} in Section~\ref{sec:bushy}).
We also use (3) as additional proof dependencies for ATP-unproved theorems when training premise selectors (Section~\ref{sec:trainprems}).

\section{Strategies and Portfolios}
\label{sec:stratsports}
\textbf{Strategies:} E, ENIGMA, Vampire and Deepire are parameterized by
  ATP strategies and their combinations.  While ENIGMA-style guidance
  typically involves the application of a larger (neural, tree-based,
  etc.) and possibly slower statistical model to the clauses, standard
  ATP strategies typically consist of much faster clause evaluation
  functions and programs written in a DSL provided by the prover. Such
  programs can again be invented and learned in various ways for
  particular classes of problems.
  For the experiments here we have used many ATP strategies invented automatically
  by the BliStr/Tune %
  systems~\cite{blistr,JakubuvU17,JakubuvSU17}.
   They implement feedback loops that interleave targeted \emph{parameter search} on problem clusters
   using engines like ParamILS~\cite{ParamILS-JAIR}, with a large-scale evaluation of the invented strategies used for evolving the problem clustering.
   Starting from few strategies, BliStr/Tune typically evolve %
   each strategy on the
   problems where the strategy performs best.
   During our experiments with the systems we have developed several thousand E
   Prover strategies, many of them targeted to Mizar problems.
   Some of these are mentioned in the experiments in Section~\ref{Results}.

\textbf{Robust Portfolios:} Larger AI/TP systems and metasystems
rely on portfolios~\cite{Tammet98} of complementary strategies that
attack the problems serially or in parallel using a global time
limit. In the presence of premise selection and multiple ATPs, such
portfolios may consist of tens to hundreds of different methods. The
larger the space of methods, the larger is the risk of overfitting the
portfolio during its construction on a particular set of problems.
For example, naive construction of ``optimal'' portfolios by using SAT
solvers for the set-cover problem (where each strategy covers some
part of the solution space) often leads to portfolios that are highly
specialized to the particular set of problems. This is mitigated in
more robust %
methods such as the \emph{greedy
  cover}, however, the overfitting there can still be significant. E.g.,
a 14-strategy greedy cover built in the Mizar40
experiments~\cite{KaliszykU13b} solved \qty{44.1}{\percent} of the random subset used for its
construction, while it solved only \qty{40.6}{\percent} of the whole MML,
i.e., \qty{8}{\percent} less.

To improve on this, we propose a more robust way of portfolio
construction here, based again on the machine-learning ideas of
controlling overfitting. Instead of simply constructing one greedy
cover $C$ (with a certain time budget) on the whole development set
$D$ and evaluating it in the holdout set $H$, we first split $D$
randomly into two equal size halves $D_1$ and $D_2$. Then we construct a
greedy cover $C_1$ only on $D_1$, and evaluate its performance also on
$D_2$ and the full set $D$. This is repeated $n$ times (we use
$n=1000$), which for large enough $n$ typically guarantees that the
greedy cover $C_1^i$ will for some of the random splits $D_1^i$, $D_2^i$
overfit very little (or even underfit). This can be further improved
by evaluating the best (strongest and least overfitting) covers on
many other random splits and selecting the most robust ones. We use
this in Section~\ref{Results} to build a portfolio that performs only
\qty{3.5}{\percent} worse on the (unseen) holdout set than on the development set used for its construction.

\section{Experiments and Results}
\label{Results}

\subsection{Bushy Experiments and Timeline}
\label{sec:bushy}

The final list of all \num{43717} Mizar problems proved by ATPs in our evaluation is available on our web
page.\footnoteA{\url{http://grid01.ciirc.cvut.cz/~mptp/00proved_20210902}}
The approximate timeline of the methods and the added solutions is shown in Table~\ref{t:timeline}.
This was
continuously recorded on our web page,\footnoteA{\url{https://github.com/ai4reason/ATP_Proofs}}
which also gives an idea of how the experiments progressed and how increasingly hard problems were proved.

The large evaluation started in April 2020, as a follow-up to our work
on ENIGMA Anonymous~\cite{JakubuvCOP0U20}. By combining the methods
developed there and running with higher time limits, the number of
problems proved by ENIGMA in the bushy setting reached
\qty{65.65}{\percent} in June 2020. This was continued by iterating
the learning and proving in a large Malarea-style feedback loop. The
growing body of proofs was continuously used for training the graph
neural networks and gradient boosted guidance, which were used for
further proof attempts, combined with different search parameters and
later used also for training premise selection.

This included
many grid searches on a small random subset of the problems over the
thousands of differently trained GNNs and GBDTs corresponding to the
training epochs, and then evaluating the strongest and most
complementary ENIGMAs using the differently trained GNNs and GBDTs on
all, or just \emph{hard} (the so far ATP-unproved), problems.  The
total number of the saved snapshots of the GNNs corresponding to the
training epochs and usable for the grid searches and full evaluations
reached \num{15920} by the end of the experiments in September
2021.\footnoteA{For the grid searches, this was compounded by further
  parameters of the ENIGMA and E strategies.}  The longest GNN
training we did involved 964 epochs and 12 days on a high-end NVIDIA
V100 GPU card.\footnoteA{We generally use the same GNN hyper-parameters
  as in~\cite{OlsakKU20,JakubuvCOP0U20}
  with the exception of the number of \emph{layers} that varied here
  between \num{5} and \num{12}, providing tradeoffs between the GNN's speed and precision.}
The GNN training occasionally (but rarely) diverged after hundreds of epochs, which we handled by restarts.

The total number of proofs that we trained the ENIGMA guidance on
eventually reached more than three million, which in a pickled and
compressed form take over \qty{200}{\giga\byte}. Since the full data
do not fit into the main memory of even large servers equipped for
efficient GPU-based neural training, we have programmed custom
pipelines that continuously load, mix and unload smaller chunks of
data used for the ENIGMA training. For many problems, we obtained
hundreds of different proofs, while for some problems we may have only
a single proof. This motivated further experiments on how and with
what frequency the different proofs should be represented in the
training data. This was a part of the larger task of \emph{training data
  normalization}, which included, e.g., removing or pruning very large
proof searches in the training data that would cause memory-based GPU
crashes.

\begin{table}[t]
  \begin{tabular}{lclcl}\toprule
\emph{solved} & [\%] & \emph{date} & \emph{premises} & \emph{methods/notes }\\\midrule
    38k  & 65.65 & Jun 2020  & \SYMpremselBushy &   ENIGMA, reported on July 2nd at IJCAR'20\footnoteA{\url{https://youtu.be/XojOEpZfH4Y?t=673}}   \\
  \num{40268} & 69.57 & Oct 2020 & \SYMpremselBushy & ENIGMA \\
  \num{40994} & 70.83 & Nov 12    & \SYMpremselMin &  ENIGMA, heuristic premise minimization \\
  \num{41169} & 71.13 & Nov 12    & \SYMpremselMin &  Vampire with \qty{300}{\second} limit adds 175 \\
  \num{41792} & 72.20 & Nov 27    & \SYMpremselMin &  E/ENIGMA/Vampire with more premise minimization \\
  \num{42206} & 72.92 & Dec 7     & \SYMpremselMin & E/ENIGMA/Vampire with more premise minimization \\
  \num{42471} & 73.38 & Jan 6     & \SYMpremselGnn, \SYMpremselMix & E with BliStr/Tune strategies on \SYMpremselGnn, \SYMpremselMix{} premises \\
  \num{42519} & 73.46 & Jan 10    & many &  ENIGMA runs on all training predictions \\
  \num{42826} & 73.99 & May 14    & \SYMpremselGnn,\SYMpremselLgb,\SYMpremselKnn &  Vampire/Deepire runs -- FroCoS'21~\cite{DBLP:conf/frocos/Suda21} \\
  \num{43414} & 75.01 & Jul 26    & \SYMpremselMin,\SYMpremselBushy &  2,3-phase ENIGMA, leapfrogging  \\
  \num{43524} & 75.20 & Aug 21    & \SYMpremselMin &  3-phase ENIGMA, shifting context, leapfrog., fwd subsump.
\\ \num{43599} & 75.33 & Aug 26    & \SYMpremselLgb &  3-phase ENIGMA, leapfrogging, fwd. subsumption
\\ \num{43717} & 75.53 & Sep 2     & \SYMpremselMin & mainly Vampire/Deepire
\\\bottomrule
\end{tabular}
\caption{Timeline of the experiments. \SYMpremselBushy{} are standard bushy premises, \SYMpremselMin{}  are subproblem-minimized premises, \SYMpremselGnn, \SYMpremselLgb, and \SYMpremselKnn{}  are GNN/LightGDB/kNN-based premises, and \SYMpremselMix{} their ensembles.}\label{t:timeline}
\end{table}

The \SI{75}{\percent} milestone was reached on July 26
2021\footnoteA{\url{https://github.com/ai4reason/ATP_Proofs/blob/master/75percent_announce.md}}
by using the freshly developed %
2 and 3-phase ENIGMAs,
together with differently parameterized leapfrogging (Section~\ref{s:related}) runs.  The
strongest single 3-phase ENIGMA strategy has reached \SI{56.4}{\percent}
performance in \qty{30}{\second} on the bushy problems when trained and evaluated
in a rigorous train/dev/holdout setting~\cite{GoertzelCJOU21}. This
best ENIGMA uses a parental threshold of $0.01$, 2-phase threshold of
$0.1$, and context and query sizes of $768$ and $256$. Its
(server-based) GNN has 10 layers trained on at most three proofs for
each problem in the training set. See also Section~\ref{s:mml1382} for
its evaluation on a set of completely new \num{13 370} problems in \num{242} new articles of a
later version of MML.

\subsection{Training Data for Premise Selection}
\label{sec:trainprems}

After several months of running the learning/proving loop in various
ways on the problems, we used the collected data for training premise
selection methods. In particular, at that point, there were \num{41504}
ATP-proved problems for which we typically had many alternative proofs
and sets of premises, yielding \num{621642} unique ATP proof dependencies.
Since in the hammering scenarios we can also analyze
the human-written proofs and learn from them, we have added for each ATP-unproved problem
$P$ its premises obtained by taking the union of the ATP dependencies
of all subproblems of $P$. In other words, we use subproblem-based
premise minimization (Section~\ref{subproofs}) for the remaining hard problems. This adds \num{16651} examples to
the premise selection dataset. This dataset of \num{638293} unique proof
dependencies is then used in various ways for training and evaluating
the premise selection methods on MML. In comparison with the Mizar40
experiments this is about six times more proof data.
As usual in machine learning experiments, we also split the whole set of Mizar problems into the \emph{train}ing, \emph{devel}opment, and \emph{holdout} subsets, using a 90 : 5 : 5 ratio.
This yields $\num{52125}$ problems in the training set, $\num{2896}$ in devel, and $\num{2896}$ in the holdout set.

\subsection{Training the Premise Selectors} %

We first train kNN and naive Bayes in multiple ways on the training subset using the
different features (Section~\ref{sec:knn-nb}) and their combinations.
For training the GNN and LightGBM, we first use kNN-based
pre-selection to choose 512 most relevant premises for each
problem. When training, we add for each example its positives (the
real dependencies) and subtract them from the 512 premises
pre-selected by kNN, thus forming the set of the negatives for the
example. The GNN and LightGBM are thus trained to correct the mistakes
done by kNN (a form of \emph{boosting}). When predicting, this is done
in the same way, i.e., first we use the trained kNN to preselect 512
premises which are then ranked by the GNN/LightGBM. We use both score
thresholds (e.g., including all premises with score better than $0$,
$-1$ or $-3$), and fixed-sized slices as in other premise selection
methods. With the same best version of ENIGMA, the strongest GNN-based
predictor (\premselGnn{-1}) solves 1089 problems compared to 870 solved when using the baseline kNN,
which is a large (\qty{25.2}{\percent}) improvement. The GNN also
outperforms LightGBM, which seems to overfit more easily on the
training data. Table~\ref{t:mleval} shows the detailed performance
on the devel and holdout sets
of the main methods used in the evaluation.
\begin{table}[tb]
  \centering
  \begin{tabular}{lcccccccccc}\toprule
    Model & \multicolumn{2}{c}{100-Cover} & \multicolumn{2}{c}{100-Prec} & \multicolumn{2}{c}{Recall} & \multicolumn{2}{c}{AUC} & \multicolumn{2}{c}{Avg. Rank} \\
          & D & H & D & H & D & H & D & H & D & H \\ \midrule
\premselKnn{cp}{var} & 83.3 & 82.3 & 8.837 & 8.713 & 386.8 & 401.9 & 92.03 & 91.27 & 90.17 & 97.98 \\
\premselKnn{au}{var} & 83.5 & 82.7 & 8.855 & 8.754 & 383.32 & 401.54 & 92.13 & 91.36 & 89.21 & 97.19 \\
\premselKnn{mi}{var} & 82.6 & 81.8 & 8.700 & 8.596 & 401.89 & 418.40 & 91.32 & 90.59 & 97.30 & 104.88 \\
\premselKnn{s0}{var} & 83.6 & 82.9 & 8.851 & 8.785 & 382.31 & 399.10 & 92.19 & 91.39 & 88.53 & 96.84 \\
\premselNb{cp} & 87.8 & 87.0 & 9.739 & 9.665 & 300.49 & 310.72 & 94.77 & 94.32 & 62.64 & 67.51 \\
\premselNb{au} & 88.0 & 87.5 & 9.748 & 9.714 & 298.66 & 307.82 & 94.84 & 94.44 & 61.99 & 66.31 \\
\premselNb{mi} & 83.3 & 83.5 & 9.358 & 9.367 & 382.87 & 379.24 & 92.53 & 92.41 & 85.39 & 86.88 \\
\premselNb{s0} & 88.3 & 87.5 & 9.776 & 9.720 & 299.10 & 308.67 & 94.85 & 94.41 & 61.85 & 66.60 \\
\premselNb{mi,chrono} & 83.9 & 82.7 & 9.151 & 9.010 & 384.68 & 393.37 & 92.33 & 91.75 & 87.23 & 93.27 \\
\premselLgb{} & 82.9 & 83.1 & 9.077 & 9.090 & 410.06 & 408.06 & 91.53 & 91.26 & 95.11 & 97.74 \\
\premselGnn{} & 87.4 & 86.3 & 9.408 & 9.282 & 241.22 & 249.32 & 88.45 & 87.42 & 66.69 & 71.87 \\ \midrule
\premselMix{\SYMpremselNb,\SYMpremselKnn}{.5,.5,\&\textsf{avg}} & 87.8 & 87.1 & 9.606 & 9.522 & 291.27 & 304.43 & 95.06 & 94.53 & 59.81 & 65.47 \\
\premselMix{\SYMpremselNb,\SYMpremselKnn}{.5,.5,\&\textsf{geo}} & 89.4 & 88.7 & 9.806 & 9.733 & 277.75 & 288.53 & 95.53 & 95.04 & 55.13 & 60.27 \\
\premselMix{\SYMpremselNb,\SYMpremselKnn}{.5,.5,\&\textsf{har}} & 89.4 & 88.9 & 9.822 & 9.780 & 276.34 & 286.23 & 95.53 & 95.08 & 55.06 & 59.94 \\
\premselMix{\SYMpremselNb,\SYMpremselKnn}{.5,.5,\&\textsf{min}} & 89.0 & 88.4 & 9.753 & 9.707 & 279.88 & 289.41 & 95.37 & 94.95 & 56.70 & 61.19 \\
\premselMix{\SYMpremselNb,\SYMpremselGnn,\SYMpremselKnn}{.5,.25,.25} & 92.1 & 91.1 & 10.237 & 10.160 & 228.79 & 248.16 & 96.64 & 96.20 & 44.06 & 48.76 \\
\premselMix{\SYMpremselNb,\SYMpremselKnn,\SYMpremselLgb,\SYMpremselGnn}{.5,.2,.2,.1} & 92.5 & 91.5 & 10.297 & 10.219 & 210.31 & 227.74 & 96.93 & 96.56 & 41.20 & 45.17 \\
\premselMix{\SYMpremselNb,\SYMpremselGnn,\SYMpremselKnn}{.33,.33,.33} & 91.3 & 90.4 & 10.091 & 10.014 & 261.10 & 272.85 & 96.20 & 95.71 & 48.51 & 53.64 \\ \bottomrule
  \end{tabular}
  \caption{Machine learning evaluation of the premise selection models on the \textbf{D}evelopment and \textbf{H}oldout datasets. Note that the evaluation of GNN is presented here only for completeness, in practice we use it with a score-based threshold and fewer premises.}
  \label{t:mleval}
\end{table}

\subsection{ENIGMA Experiments on the Premise Selection Data}

First, to train ENIGMA on the premise selection problems, we perform several prove/learn iterations with
ENIGMA/GBDT  on our premise slices.
In loop (1), we start with three selected slices \premselGnn{-1}, \premselLgb{0.1}, and \premselKnn{}{64}, which were found experimentally to be complementary.
We evaluate strategy \Str{1} (\textsf{bls0f17}) on the three slices obtaining
\num{20604} proved training problems.
We train several decision tree (GBDT) models with various learning
hyperparameters (tree leaves count, tree depth, ENIGMA features used).
We use all the training proofs
available.
In loop (2), we evaluate several ENIGMA
models trained on \SYMpremselBushy~(bushy problems) to obtain additional
training data.
After few training/evaluation iterations, the training data might start
accumulating many proofs for some (easier) problems solved by many strategies.
From loop (2) on, we, therefore, use only a limited number of proofs per problem.
We either select randomly up to $6$ proofs for each problem, or we select
only specific proofs (e.g., the shortest, longest, and one medium-length proof).
In loop (3), additional training data are added by ENIGMA/GNN runs on the premise
slices, with GNN trained on the GBDT runs.
In loop (4), we consider training data from $7$ additional slices (variants of
\SYMpremselGnn, \SYMpremselLgb, \SYMpremselKnn), obtained by running ENIGMA
models trained of bushy problems.
In loop (5), we extend the training data with bushy proofs of unsolved
training problems obtained by our various previous efforts.

\begin{table}[t]
\begin{center}
\begin{minipage}[]{0.49\textwidth}
\begin{tabular}{lcccc}
   \toprule
   \emph{loop} &\emph{trains} & \emph{devel} & \multicolumn{2}{c}{\emph{devel cover} }
   \\      &       & (union)& (in \qty{420}{\second}) & [\%]
\\\hline
init & \num{20604} & 1215&  -  & - \\
(1) & $\num{25240}$ & $1601$ & $1516$ & $52.33$\\
(2) & $\num{25725}$ & $1669$ & $1555$ & $53.69$\\
(3) & $\num{25887}$ & $1679$ & $1560$ & $53.88$\\
(4) & $\num{29266}$ & $1716$ & $1591$ & $54.94$\\
(5) & $\num{37053}$ & $1735$ & $1610$ & $55.59$\\
\bottomrule
\end{tabular}
\end{minipage}
\begin{minipage}[]{0.49\textwidth}
   \begin{tabular}{llrc}
\toprule
   \emph{prover} (\qty{420}{\second})  & cover & \emph{pairs} & [\%] \\
\midrule
E 2.6 (auto-schedule)    & 1430  & $14$  & 49.38 \\
Vampire 4.0 (CASC)       & 1536  & $14$  & 53.03 \\
BliStr/Tune                   & 1582  & $210$ & 54.62 \\
ENIGMA/GBDT                   & 1610  & $42$  & 55.59 \\
ENIGMA/GNN                    & 1670  & $84$  & 57.66 \\
\bottomrule
\end{tabular}
\end{minipage}
\end{center}
\caption{Training of ENIGMA/GBDT models (left), and best covers of development set (right).}
\label{fig:enigma-train}
\end{table}

Starting from $1215$ solved development problems, we ended up with %
$1735$ problems solved after the fifth iteration.
While we train GBDT models only on few selected slices, we evaluate the models
on many more, up to $56$, development slices covering all families
$\SYMpremselGnn$, $\SYMpremselLgb$, $\SYMpremselKnn$, $\SYMpremselMix$, and
$\SYMpremselNb$.
We report the increasing number of training problems (\emph{trains}) and the
total of number of solved development problems by all the evaluated
strategy/slice pairs (\emph{devel} union).
Since every strategy/slice pair is evaluated in $10$ seconds, we construct the
greedy cover of best $42$ strategy/slice pairs, to approximate the best possible
result obtainable in \qty{420}{\second} (see columns \emph{devel cover}).
Since the development set has not been used in any way to train the GBDT
models, we can see this as an approximation of the best possible result on the
holdout set.

We reach \qty{55.59}{\percent} of problems solvable in \qty{420}{\second}, only with
ENIGMA/GBDT models.
To compare this result to other methods, we construct compatible greedy covers
for E Prover in \emph{auto-schedule mode}, and for Vampire in \emph{CASC mode},
that is, in their respective strongest default settings.
We evaluate both provers on all $56$ development slices, with \qty{30}{\second} limit
per problem.
For each prover, we construct a greedy cover of best $14$ slices, again
approximating the best possible result obtainable in \qty{420}{\second}.
BliStr/Tune is our previously invented portfolio of $15$ E strategies for Mizar
bushy problems.
We evaluate all $15$ strategies on all $56$ slices with $2$ seconds per problem.
Similarly, the greedy cover of length $210$ is constructed.
The column \emph{pairs} specifies the greedy cover length considered in
each case.
The time limit for each strategy/slice pair is $420/\mathit{pairs}$.

The training data obtained in five loops were finally used to train new
ENIGMA/GNN models for premise selection slices.
Various GNN models were trained (various numbers of layers, networks from various epochs) and evaluated with the limit of \qty{5}{\second}.
As before, we construct the greedy cover of length $84$ to simulate the best
possible run in \qty{420}{\second}.
ENIGMA/GNN performs even better then ENIGMA/GBDT, solving \qty{57.66}{\percent} problems.
The two ENIGMA/* portfolios cover together $1701$ development problems (in \qty{840}{\second}),
suggesting a decent complementarity of the methods.
Note that only the ENIGMA/GBDT strategies can cover up to $1735$ (see column
\emph{devel} on the left), which is \qty{59,9}{\percent} of the development set.

Most of our ENIGMA models are combined with the baseline strategy
$\textsf{bls0f17}$.
This together with $\textsf{bls05fc}$ are two strategies invented by
BliStr/Tune~\cite{JakubuvU18a} which perform well on premise selection
data.
We additionally use another two older BliStr~\cite{blistr} strategies
$\textsf{mzr02}$ and $\textsf{mzr03}$ which perform well on bushy problems.
We usually combine training data only from strategies with compatible term
ordering and literal selection setting.
However, data from strategies with incompatible orderings, were found useful
when used in a reasonably small amounts.
Few other BliStr and Vampire strategies, together with E in the \emph{auto} mode, are used to gather additional solved development problems.
With all our methods (ENIGMA \& BliStr/Tune) and with additional Vampire runs of
selected strategies, we have solved more than
\SI{62.7}{\percent}
development problems.
These results provide training data for the construction of the final holdout portfolio, as described in the next section.

\subsection{Final Hammer Portfolio}

With the large database of the development results of the systems run
on the premise slices, we finally construct our ultimate hammering
portfolio. For that, we use the \emph{robust portfolio construction} method described in Section~\ref{sec:stratsports}.
In particular, we randomly split the development set into two equal-sized parts, and compute the \qty{420}{\second} greedy cover
using our whole database of results on the first part.
This greedy cover is evaluated on the second part, thus measuring the overfitting.
This randomized procedure is repeated one thousand times. Then we (manually) select the 20 strongest and least overfitting portfolios and evaluate each of them on 80 more random splits, thus measuring how balanced they are on average. Typically, they reach up to \SI{60.5}{\percent} performance on the whole devel set, so we choose a threshold of \SI{59.5}{\percent} on the 160 random halves to measure the imbalance. The most balanced portfolio wins with 135 of the 160
random halves passing the threshold.

This final 420-second portfolio has 95 slices that solve 1749 (\qty{60.4}{\percent}) of the devel problems
and
1690 (\qty{58.36}{\percent}) of the holdout problems.
Table~\ref{tab:portfolio1} shows the initial segment of 13 slices of this portfolio with the numbers of problems solved.
The full portfolio is presented in Table~\ref{fig:portfolio}. The
first number \emph{t} is the number of seconds to run the slice.  The
\emph{base} column specifies the ATP strategy used, and \emph{ENIGMA}
describes what kinds of ENIGMA models are used (if any).  We can see
that GNN models dominate the schedule with fast runs.  The schedule is
closed by longer runs, notably also GBDT models, which while evaluated
in a single-CPU setting, need several seconds to load the model.  This
means that we are favoring the GNN ENIGMAs thanks to the use of the
preloaded GNN server, and a further improvement is likely if we also
preload the GBDT models.
Our single strongest GNN-based strategy solves 1178 of the holdout problems in \qty{30}{\second} using the  \premselGnn{-1} predictions.
This is \SI{39.5}{\percent}, which is only \SI{1.1}{\percent} less than the \SI{40.6}{\percent} solved by the full
\qty{420}{\second} portfolio constructed in the Mizar40 experiments.

\begin{table}[tb]
  \centerline{
  \begin{tabular}{lllllllllllll}
\toprule
  \premselGnn{} & \premselMix{}{5221} & \premselLgb{} & \premselMix{}{5221} & \premselNb{uni} & \premselMix{}{5221} & \premselNb{eni} & \premselNb{eni} & \premselMix{}{5221} & \premselGnn{} & \premselMix{}{55} & \premselMix{}{533} & \premselMix{}{533} \\
\midrule
  GNN & GNN & GNN & GNN & V & V & GNN & GNN & GNN & GNN & V & V & GNN \\
    984 & 1142 & 1215 & 1263 & 1297 & 1325 & 1346 & 1370 & 1381 & 1393 & 1405 & 1419 & 1444 \\
    1013 & 1157 & 1240 & 1275 & 1305 & 1321 & 1346 & 1364 & 1378 & 1386 & 1398 & 1407 & 1436 \\
\bottomrule
  \end{tabular}
  }
  \caption{The 13-slice prefix of the final portfolio of the 95 slices. Each column presents the premise selection method, the ATP method, and the number of problems solved up this slice cumulatively on the development and holdout sets.
     ``V'' stands for Vampire and ``GNN'' is ENIGMA/GNN model based on \textsf{bls0f17}.
   Moreover,
$\premselMix{}{5221} =
\premselMix{\SYMpremselNb,\SYMpremselKnn,\SYMpremselLgb,\SYMpremselGnn}{.5,.2,.2,.1}$
and
$\premselMix{}{55}=\premselMix{\SYMpremselNb,\SYMpremselKnn}{.5,.5,\textrm{avg}}$
and
$\premselMix{}{533}=\premselMix{\SYMpremselNb,\SYMpremselGnn,\SYMpremselKnn}{.5,.25,.25}$.
}
  \label{tab:portfolio1}
\end{table}

\begin{table}
\begin{minipage}[]{0.49\textwidth}
\begin{tabular}{lllllll}
\toprule
\emph{t} & \emph{base} & \emph{ENIGMA} & \emph{slice}  \\\midrule
 2  & \textsf{bls0f17}  & \textrm{GNN}  & \premselGnn{-1}  \\ %
 2  & \textsf{bls0f17}  & \textrm{GNN}  & \premselMix{\SYMpremselNb,\SYMpremselKnn,\SYMpremselLgb,\SYMpremselGnn}{.5,.2,.2,.1}  \\ %
 2  & \textsf{bls0f17}  & \textrm{GNN}  & \premselLgb{0.1}  \\ %
 2  & \textsf{bls0f17}  & \textrm{GNN}  & \premselMix{\SYMpremselNb,\SYMpremselKnn,\SYMpremselLgb,\SYMpremselGnn}{.5,.2,.2,.1}  \\ %
 2  & \textsf{vampire}  & \textrm{-}  & \premselNb{uni}  \\ %
 2  & \textsf{vampire}  & \textrm{-}  & \premselMix{\SYMpremselNb,\SYMpremselKnn,\SYMpremselLgb,\SYMpremselGnn}{.5,.2,.2,.1}  \\ %
 2  & \textsf{bls0f17}  & \textrm{GNN}  & \premselNb{eni}  \\ %
 2  & \textsf{bls0f17}  & \textrm{GNN}  & \premselNb{eni}  \\ %
 2  & \textsf{bls0f17}  & \textrm{GNN}  & \premselMix{\SYMpremselNb,\SYMpremselKnn,\SYMpremselLgb,\SYMpremselGnn}{.5,.2,.2,.1}  \\ %
 2  & \textsf{bls0f17}  & \textrm{GNN}  & \premselGnn{-3}  \\ %
 2  & \textsf{vampire}  & \textrm{-}  & \premselMix{\SYMpremselNb,\SYMpremselKnn}{.5,.5}  \\ %
 2  & \textsf{vampire}  & \textrm{-}  & \premselMix{\SYMpremselNb,\SYMpremselGnn,\SYMpremselKnn}{.5,.25,.25}  \\ %
 5  & \textsf{bls0f17}  & \textrm{GNN}  & \premselMix{\SYMpremselNb,\SYMpremselGnn,\SYMpremselKnn}{.5,.25,.25}  \\ %
 2  & \textsf{vampire}  & \textrm{-}  & \premselMix{\SYMpremselNb,\SYMpremselKnn}{.5,.5}  \\ %
 2  & \textsf{vampire}  & \textrm{-}  & \premselKnn{au}{var}  \\ %
 2  & \textsf{mzr02}  & \textrm{-}  & \premselKnn{cp}{var}  \\ %
 2  & \textsf{bls0f17}  & \textrm{GNN}  & \premselGnn{0 }  \\ %
 2  & \textsf{vampire-16}  & \textrm{-}  & \premselGnn{-5}  \\ %
 2  & \textsf{vampire}  & \textrm{-}  & \premselNb{uni}  \\ %
 2  & \textsf{bls0f17}  & \textrm{GNN}  & \premselNb{eni}  \\ %
 5  & \textsf{bls0f17}  & \textrm{GNN}  & \premselNb{au}  \\ %
 2  & \textsf{vampire}  & \textrm{-}  & \premselMix{\SYMpremselNb,\SYMpremselKnn}{.5,.5}  \\ %
 2  & \textsf{vampire}  & \textrm{-}  & \premselMix{\SYMpremselNb,\SYMpremselKnn}{.5,.5}  \\ %
 5  & \textsf{bls0f17}  & \textrm{GNN}  & \premselMix{\SYMpremselNb,\SYMpremselKnn}{.5,.5}  \\ %
 2  & \textsf{bls0f17}  & \textrm{GNN}  & \premselMix{\SYMpremselNb,\SYMpremselGnn,\SYMpremselKnn}{.5,.25,.25}  \\ %
 2  & \textsf{vampire-16}  & \textrm{-}  & \premselMix{\SYMpremselNb,\SYMpremselKnn,\SYMpremselLgb,\SYMpremselGnn}{.5,.2,.2,.1}  \\ %
 5  & \textsf{bls0f17}  & \textrm{GNN}  & \premselLgb{0.05}  \\ %
 2  & \textsf{vampire-18}  & \textrm{-}  & \premselGnn{}  \\ %
 2  & \textsf{mzr22}  & \textrm{-}  & \premselMix{\SYMpremselNb,\SYMpremselKnn,\SYMpremselLgb}{.5,.2,.3}  \\ %
 2  & \textsf{vampire}  & \textrm{-}  & \premselMix{\SYMpremselNb,\SYMpremselKnn}{.5,.5,geo} \\ %
 2  & \textsf{vampire}  & \textrm{-}  & \premselMix{\SYMpremselNb,\SYMpremselKnn}{.5,.5}  \\ %
 2  & \textsf{vampire}  & \textrm{-}  & \premselKnn{cp}{var}  \\ %
 2  & \textsf{BliStr-edc9}  & \textrm{-}  & \premselMix{\SYMpremselNb,\SYMpremselKnn,\SYMpremselLgb}{.5,.2,.3}  \\ %
 2  & \textsf{vampire}  & \textrm{-}  & \premselMix{\SYMpremselNb,\SYMpremselKnn}{.5,.5,chrono}  \\ %
 2  & \textsf{vampire}  & \textrm{-}  & \premselNb{uni}  \\ %
 5  & \textsf{bls0f17}  & \textrm{GNN}  & \premselMix{\SYMpremselNb,\SYMpremselGnn,\SYMpremselKnn}{.5,.25,.25}  \\ %
10  & \textsf{mzr02}  & \textrm{GNN}  & \premselLgb{0}  \\ %
 5  & \textsf{bls0f17}  & \textrm{GNN}  & \premselMix{\SYMpremselNb,\SYMpremselKnn,\SYMpremselLgb,\SYMpremselGnn}{.5,.2,.2,.1}  \\ %
 5  & \textsf{bls0f17}  & \textrm{GNN}  & \premselNb{au}  \\ %
 2  & \textsf{bls0f17}  & \textrm{GNN}  & \premselNb{eni}  \\ %
 2  & \textsf{vampire-2}  & \textrm{-}  & \premselGnn{16}  \\ %
 2  & \textsf{vampire-16}  & \textrm{-}  & \premselNb{au}  \\ %
 2  & \textsf{vampire}  & \textrm{-}  & \premselNb{au}  \\ %
 2  & \textsf{E-auto}  & \textrm{-}  & \premselLgb{96}  \\ %
10  & \textsf{bls05fc} & \textrm{GBDT}  & \premselLgb{0.25}  \\ %
 2  & \textsf{vampire}  & \textrm{-}  & \premselMix{\SYMpremselNb,\SYMpremselKnn}{.5,.5,min}  \\ %
 2  & \textsf{vampire-16}  & \textrm{-}  & \premselNb{sub}  \\ %
 2  & \textsf{vampire-16}  & \textrm{-}  & \premselMix{\SYMpremselNb,\SYMpremselGnn,\SYMpremselKnn}{.5,.25,.25}  \\ %
\bottomrule
\end{tabular}
\end{minipage}
\begin{minipage}[]{0.49\textwidth}
\begin{tabular}{lllllll}
\toprule
\emph{t} & \emph{base} & \emph{ENIGMA} & \emph{slice}  \\\midrule
 2  & \textsf{vampire}  & \textrm{-}  & \premselMix{\SYMpremselNb,\SYMpremselGnn,\SYMpremselKnn}{.5,.25,.25}  \\ %
 2  & \textsf{Blistr-5fce}  & \textrm{-}  & \premselMix{\SYMpremselNb,\SYMpremselKnn,\SYMpremselLgb}{.5,.2,.3}  \\ %
 2  & \textsf{vampire}  & \textrm{-}  & \premselMix{\SYMpremselNb,\SYMpremselKnn}{.5,.5}  \\ %
 2  & \textsf{E-auto}  & \textrm{-}  & \premselGnn{-2}  \\ %
 2  & \textsf{vampire}  & \textrm{-}  & \premselKnn{uni}{var}  \\ %
 2  & \textsf{vampire-16}  & \textrm{-}  & \premselMix{\SYMpremselNb,\SYMpremselKnn}{.5,.5}  \\ %
 2  & \textsf{vampire}  & \textrm{-}  & \premselMix{\SYMpremselNb,\SYMpremselKnn}{.5,.5,geo}  \\ %
 2  & \textsf{vampire}  & \textrm{-}  & \premselKnn{uni}{var}  \\ %
 2  & \textsf{vampire-21}  & \textrm{-}  & \premselGnn{}  \\ %
 2  & \textsf{vampire}  & \textrm{-}  & \premselNb{sub}  \\ %
 2  & \textsf{vampire}  & \textrm{-}  & \premselNb{uni}  \\ %
 2  & \textsf{E-auto}  & \textrm{-}  & \premselMix{\SYMpremselNb,\SYMpremselGnn,\SYMpremselKnn}{.5,.25,.25}  \\ %
 2  & \textsf{vampire}  & \textrm{-}  & \premselMix{\SYMpremselNb,\SYMpremselKnn}{.5,.5,min}  \\ %
 2  & \textsf{vampire}  & \textrm{-}  & \premselNb{cp}  \\ %
 2  & \textsf{bls0f17}  & \textrm{GNN}  & \premselKnn{eni}{var}  \\ %
 5  & \textsf{bls0f17}  & \textrm{GNN}  & \premselLgb{0.01}  \\ %
10  & \textsf{bls05fc}  & \textrm{GBDT}  & \premselMix{\SYMpremselNb,\SYMpremselGnn,\SYMpremselKnn}{.33,.33,.33}  \\ %
 5  & \textsf{bls0f17}  & \textrm{GNN}  & \premselMix{\SYMpremselNb,\SYMpremselKnn}{.5,.5}  \\ %
 5  & \textsf{bls0f17}  & \textrm{GNN}  & \premselNb{eni}  \\ %
 5  & \textsf{bls0f17}  & \textrm{GNN}  & \premselMix{\SYMpremselNb,\SYMpremselGnn,\SYMpremselKnn}{.5,.25,.25}  \\ %
10  & \textsf{bls0f17}  & \textrm{GNN}  & \premselMix{\SYMpremselNb,\SYMpremselKnn,\SYMpremselLgb,\SYMpremselGnn}{.5,.2,.2,.1}  \\ %
10  & \textsf{mzr03}  & \textrm{GBDT}  & \premselGnn{64}  \\ %
10  & \textsf{bls05fc}  & \textrm{GBDT}  & \premselMix{\SYMpremselNb,\SYMpremselGnn,\SYMpremselKnn}{.33,.33,.33}  \\ %
10  & \textsf{mzr02} & \textrm{GBDT}   & \premselKnn{}{short}  \\ %
10  & \textsf{bls0f17}  & \textrm{GNN}  & \premselMix{\SYMpremselNb,\SYMpremselGnn,\SYMpremselKnn}{.5,.25,.25}  \\ %
10  & \textsf{bls0f17}  & \textrm{GNN}  & \premselMix{\SYMpremselNb,\SYMpremselKnn,\SYMpremselLgb,\SYMpremselGnn}{.5,.2,.2,.1}  \\ %
10  & \textsf{bls0f17}  & \textrm{GBDT}  & \premselLgb{0.01}  \\ %
 5  & \textsf{bls0f17}  & \textrm{GNN}  & \premselNb{au}  \\ %
 5  & \textsf{bls0f17}  & \textrm{GNN}  & \premselLgb{0.005}  \\ %
10  & \textsf{bls0f17}  & \textrm{GNN}  & \premselMix{\SYMpremselNb,\SYMpremselKnn,\SYMpremselLgb,\SYMpremselGnn}{.5,.2,.2,.1}  \\ %
 5  & \textsf{bls0f17}  & \textrm{GNN}  & \premselLgb{0.01}  \\ %
 5  & \textsf{bls0f17}  & \textrm{GNN}  & \premselNb{au}  \\ %
 5  & \textsf{bls0f17}  & \textrm{GNN}  & \premselMix{\SYMpremselNb,\SYMpremselKnn}{.5,.5}  \\ %
 5  & \textsf{mzr03}  & \textrm{-}  & \premselMix{\SYMpremselNb,\SYMpremselKnn,\SYMpremselLgb,\SYMpremselGnn}{.25,.25,.25,.25}  \\ %
10  & \textsf{bls0f17}  & \textrm{GNN}  & \premselGnn{-1}  \\ %
 5  & \textsf{bls0f17}  & \textrm{GNN}  & \premselGnn{0 }  \\ %
10  & \textsf{bls05fc}  & \textrm{GBDT}  & \premselKnn{}{short}  \\ %
 5  & \textsf{bls0f17}  & \textrm{GNN}  & \premselNb{au}  \\ %
 5  & \textsf{bls0f17}  & \textrm{GNN}  & \premselMix{\SYMpremselNb,\SYMpremselKnn}{.5,.5}  \\ %
10  & \textsf{bls0f17}  & \textrm{GNN}  & \premselGnn{0.5}  \\ %
10  & \textsf{bls0f17}  & \textrm{GNN}  & \premselNb{eni}  \\ %
10  & \textsf{bls0f17}  & \textrm{GBDT}  & \premselLgb{0.01}  \\ %
10  & \textsf{bls0f17}  & \textrm{GBDT}  & \premselMix{\SYMpremselNb,\SYMpremselKnn}{.5,.5, \& \textsf{min}}  \\ %
10  & \textsf{bls0f17}  & \textrm{GNN}  & \premselNb{au}  \\ %
10  & \textsf{bls0f17}  & \textrm{GNN}  & \premselMix{\SYMpremselNb,\SYMpremselKnn}{.5,.5}  \\ %
10  & \textsf{bls0f17}  & \textrm{GNN}  & \premselNb{au}  \\ %
10  & \textsf{mzr03}  & \textrm{GBDT}  & \premselMix{\SYMpremselNb,\SYMpremselGnn,\SYMpremselKnn}{.33,.33,.33}  \\ %
\bottomrule
\end{tabular}
\end{minipage}
\caption{The final Mizar hammer portfolio for \qty{420}{\second}.}
\label{fig:portfolio}
\end{table}

\subsection{Transfer to MML 1382}\label{s:mml1382}

In the final experiment, we run for \qty{120}{\second} the best trained ENIGMA (3-phase, see Section~\ref{sec:bushy})
on the bushy problems from a new version of Mizar (1382) that has 242 new
articles and \num{13370} theorems in them. ENIGMA not only never trained on any of these articles, but also never saw the new terminology introduced there.
We also run the standard E auto-schedule for \qty{120}{\second} on the new version. ENIGMA proves \num{37094} (\qty{52.7}{\percent}) of the \num{70396} problems in the new library, while the E auto-schedule proves \num{24158} (\qty{34.32}{\percent}) of them. ENIGMA thus improves over E by \qty{53.55}{\percent} on the new library.
We compare this with the old MML, where the trained ENIGMA solves \num{34528} (\qty{59.65}{\percent}) of the \num{57880} problems, and E solves \num{22119} (\qty{38.22}{\percent}), i.e.,
the relative improvement there is \qty{56.10}{\percent}.

Surprisingly, just on the new \num{13370} theorems -- more than half of which contain new terminology --
the ratio of ENIGMA-proved to E-proved problems is 5934 to 3751, i.e., ENIGMA is here better than E by \qty{58.20}{\percent}.
These numbers show that the performance of our
  \emph{anonymous}~\cite{JakubuvCOP0U20} logic-aware ML methods, which learn only from
  the structure of mathematical problems, is practically untouched by the transfer to the new setting with many new concepts and lemmas.
  This is quite unusual in
  today's machine learning which seems dominated by large language models that
  typically struggle on new
  terminology.

\section{Proofs}
\label{Proofs}
As the main experiments progressed from spring 2020 to summer 2021,
we have collected interesting examples of automatically found
proofs and published their summary descriptions on our web
page.\footnoteA{\mbox{\url{https://github.com/ai4reason/ATP_Proofs}}}
As of September 2021 there were over 200 of such
example proofs, initially with ATP length in tens of clause
steps, and gradually reaching hundreds of clause
steps. Initially these were proofs found in the bushy setting,
with proofs done in the chainy (premise-selection) setting added
later, typically to show the effect of alternative
premises.

One of the earliest proofs that we put on the web
page\footnoteA{\url{https://bit.ly/3Spmf26}}
is
\th{NEWTON:72}\footnoteA{\url{https://bit.ly/3ILEkEp}}
proving that for every natural number there exists a larger prime:

\begin{footnotesize}
\begin{verbatim}
for l being Nat ex p being Prime st p is prime & p > l
\end{verbatim}
\end{footnotesize}

\noindent The
ENIGMA
proof\footnoteA{\url{https://bit.ly/3Z2iXo3}}
starts from 328 preselected Mizar facts which translate to 549 initial
clauses. The search is guided by a particular version of the GNN
running at that time (April 2020) on the CPU. Since this is
relatively costly, the proof search generated only 2856 nontrivial
clauses in \qty{6}{\second}, doing 734 nontrivial given clause loops. The final
proof takes 83 clausal steps, and uses 38 of the 328 initially
provided steps. Many of them replay the arithmetical arguments done in
Mizar. An interesting point is that the guided prover is here capable
of synthesizing a nontrivial witness ($n! + 1$) by using the supplied
facts, after which the proof likely becomes reasonably
straightforward  given the knowledge in the library (see the Appendix for a more detailed discussion of this example).
In general, using
the supplied facts together with the trained learner for guided
synthesis of nontrivial witnesses seems to be one of the main
improvements brought by the ENIGMA guidance that contributed to the
new proofs in comparison with the Mizar40 evaluation. This led us to
start research of neural synthesis
of witnesses and conjectures for AI/TP settings~\cite{UrbanJ20,DBLP:conf/lpar/Gauthier20,abs-2202-11908,abs-2301-11479}.

Arithmetical reasoning, and other kinds of ``routine computation''
in general, have turned out to be areas where ENIGMA often gradually
improved by solving increasingly hard Mizar
problems and learning from them.  Such problems include reasoning about trigonometric
functions, integrals, derivatives, matrix manipulation, etc. From the
more advanced results done by 3-phase ENIGMA, this is, e.g., a 619-long proof of
\th{SINCOS10:86}\footnoteA{\url{https://bit.ly/3StOHzV}}
found in \qty{60}{\second}, doing a lot of computation about the domain and range of
arcsec,\footnoteA{\url{https://bit.ly/2YZ0OgX}}
and a 326-long proof of
\th{FDIFF_8:14},\footnoteA{\url{https://bit.ly/3IuYHV0}}
found in \qty{31}{\second}, about the derivative of
tan (ln
x).\footnoteA{\url{https://bit.ly/3SdZjTq}}

\begin{footnotesize}
\begin{verbatim}
for x being set st x in [.(- (sqrt 2)),(- 1).] holds arcsec2 . x in [.((3 / 4) * PI),PI.]

for Z being open Subset of REAL st Z c= dom (tan * ln) holds tan * ln is_differentiable_on Z
& for x being Real st x in Z holds ((tan * ln) `| Z) . x = 1 / (x * (cos . (ln . x))^2)
\end{verbatim}
\end{footnotesize}
\noindent The first proof uses 83 Mizar facts,
starting with 1025 preselected ones. Its proof search took 5344 nontrivial
given clauses and generated over 100k nontrivial clauses in total,
making the 3-phase filtering and the use of the GPU server essential
for finding the proof efficiently.
The second proof uses 55 Mizar facts, 3136 given clause loops and it generated 26.6k
nontrivial clauses.  The reader can see on our web page that there are
many solved problems of such ``mostly computational'' kind,
suggesting that such learning approaches may be suitable for
automatically gaining competence in routine computational tasks, without the need to manually program them as done, e.g., in SMT solvers.
This has motivated our research in learning reasoning
components~\cite{DBLP:conf/tableaux/ChvalovskyJOU21}.
Two less ``computational'' but still very long ATP proofs found by 3-phase ENIGMA are
\th{BORSUK_5:31}\footnoteA{\url{https://bit.ly/3KzuPJY}}
saying that the closure of rationals on (a,b) is
[a,b],\footnoteA{\url{https://bit.ly/3C0Lwa8}}
and
\th{IDEAL_1:22}\footnoteA{\url{https://bit.ly/3Z7UPQC}}
saying that commutative rings are fields iff ideals are
trivial:\footnoteA{\url{https://bit.ly/3BWqR6K}}

\begin{footnotesize}
\begin{verbatim}
for A being Subset of R^1 for a, b being real number
st a < b & A = RAT (a,b) holds Cl A = [.a,b.]

for R being non degenerated comRing holds R is Field iff
for I being Ideal of R holds I = {(0. R)} or I = the carrier of R
\end{verbatim}
\end{footnotesize}

\noindent The Mizar proof of \th{BORSUK_5:31} takes 80 lines. ENIGMA finds a
proof from 38 Mizar facts that uses 359 clausal steps in 4883 given
clause loops. On the 400k generated clauses, the multi-phase ENIGMA mechanisms work as follows. \num{133869} clauses are
frozen by parental guidance, \num{83871} are then filtered by aggressive subsumption, and \num{64364} by the first-stage LightGBM model.
\num{125489} remaining ``good'' clauses are gradually
evaluated (in 176 batched calls) by the GNN server, using a context of 1536 processed clauses.
The ENIGMA proof of \th{IDEAL_1:22} uses 48 Mizar facts and takes 493 clausal steps in 4481 given clause loops.

One example of an ATP proof made possible thanks to the premise
selector noticing alternative lemmas in the library is
\th{FIB_NUM2:69}.\footnoteA{\url{https://bit.ly/3YWIfE6}}
This theorem, called in the MML ``Carmichael's Theorem on Prime Divisors'', states that
if $m$ divides the $n$-th Fibonacci number (Fib $n$), then $m$ does
not divide any smaller Fibonacci number, provided $m, n$ are prime
numbers.\footnoteA{\url{https://bit.ly/3oGBdRz}}
The Mizar proof has 122 lines, uses induction and we cannot so far
replay it with ATPs. The premise selector, however, finds a prior library
lemma \th{FIB_NUM:5}\footnoteA{\url{https://bit.ly/3ExtvmS}}
saying that \texttt{(Fib m) gcd (Fib n) = Fib (m gcd n)}, from which
the proof follows, using 159 clausal steps, 4214
given clause loops and 32 Mizar facts.
Finally, an example of a long Deepire proof\footnoteA{\url{https://bit.ly/3klDrJr}}
using a high time limit is
\th{ORDINAL5:36},\footnoteA{\url{https://bit.ly/3SrPyRN}}
i.e.,
the $\epsilon_0= \omega^{\omega^{\omega^{...}}}$
formula for the zeroth epsilon ordinal:\footnoteA{\url{https://bit.ly/3SozGPM}}

\begin{footnotesize}
\begin{verbatim}
first_epsilon_greater_than 0 = omega |^|^ omega
\end{verbatim}
\end{footnotesize}

\noindent The search took \num{38065} given clause loops and \qty{504}{\second}. The proof has 1193 clausal steps, using 49 Mizar facts. Deepire's very efficient neural guidance took only \qty{18}{\second} of the total time here.

\section{Conclusion:  AI/TP Bet Completed} %
\label{Future}

In 2014, after the \qty{40}{\percent} numbers were obtained by Kaliszyk and Urban
both on the Flyspeck and Mizar corpora, the last author publicly
announced three AI/TP
bets\footnoteA{\url{http://ai4reason.org/aichallenges.html}} in a talk
at Institut Henri Poincare and offered to bet up to \num{10000} EUR on
them. Part of the second bet said that by 2024, \qty{60}{\percent} of the MML and
Flyspeck toplevel theorems will be provable automatically when using
the same setting as in 2014. In the HOL setting, this was done as
early as 2017/18 by the TacticToe system, which achieved \qty{66.4}{\percent} on the
HOL library in \qty{60}{\second} and \qty{69}{\percent} in \qty{120}{\second}~\cite{GauthierKU17,GauthierKUKN21}. One could however argue
that TacticToe introduced a new kind of ML-guided tactical prover that
considerably benefits from targeted, expert-written procedures
tailored to the corpora. This in particular showed in the
large boost on HOL problems that required induction, on which standard
higher-order ATPs traditionally struggled.

In this work, we largely
completed this part of the second AI/TP bet also for the Mizar library. The main caveat is
our use of more modern hardware, in particular many ENIGMAs using the
GPU server for clause evaluation. It is however clear (both from
the LightGBM experiments and from the very efficient and CPU-based
Deepire experiments) that this is not a major issue. While it is today
typically easier to use dedicated %
hardware in ML-based
experiments, there is also growing research in the extraction of faster
predictors from those trained on GPUs that can run more efficiently on
standard hardware.

\section{Acknowledgments}
The development of ENIGMA, premise selection and other methods used
here, as well as the large-scale experiments, benefited from many
informal discussions which involved (at least) Lasse Blaauwbroek, Chad
Brown, Thibault Gauthier, Mikolas Janota, Jelle Piepenbrock, Stanislaw
Purgal, Bob Veroff, and Jiri Vyskocil.  The funding for the multi-year
development of the methods and for the experiments was partially
provided by the ERC Consolidator grant \emph{AI4REASON} no.~649043
(ZG, JJ, BP, MS and JU), the European Regional Development Fund under
the Czech project AI\&Reasoning no. CZ.02.1.01/0.0/0.0/15\_003/0000466
(KC, ZG, JJ, MO, JU), the ERC Starting Grant \emph{SMART} no.~714034 (JJ,
CK, MO), the Czech Science Foundation project 20-06390Y and project
RICAIP no. 857306 under the EU-H2020 programme (MS), ERC-CZ project
POSTMAN no. LL1902 (JJ, BP), Amazon Research Awards (JU) and the EU ICT-48
2020 project TAILOR no. 952215 (JU).

\bibliography{ate11,stsbib,local}

\begin{thebibliography}{10}

\bibitem{tensorflow2015-whitepaper}
Mart\'{\i}n Abadi, Ashish Agarwal, Paul Barham, Eugene Brevdo, Zhifeng Chen,
  Craig Citro, Greg~S. Corrado, Andy Davis, Jeffrey Dean, Matthieu Devin,
  Sanjay Ghemawat, Ian Goodfellow, Andrew Harp, Geoffrey Irving, Michael Isard,
  Yangqing Jia, Rafal Jozefowicz, Lukasz Kaiser, Manjunath Kudlur, Josh
  Levenberg, Dandelion Man\'{e}, Rajat Monga, Sherry Moore, Derek Murray, Chris
  Olah, Mike Schuster, Jonathon Shlens, Benoit Steiner, Ilya Sutskever, Kunal
  Talwar, Paul Tucker, Vincent Vanhoucke, Vijay Vasudevan, Fernanda Vi\'{e}gas,
  Oriol Vinyals, Pete Warden, Martin Wattenberg, Martin Wicke, Yuan Yu, and
  Xiaoqiang Zheng.
\newblock {TensorFlow}: Large-scale machine learning on heterogeneous systems,
  2015.
\newblock Software available from tensorflow.org.
\newblock URL: \url{https://www.tensorflow.org/}.

\bibitem{abs-1108-3446}
Jesse Alama, Tom Heskes, Daniel K\"{u}hlwein, Evgeni Tsivtsivadze, and Josef
  Urban.
\newblock Premise selection for mathematics by corpus analysis and kernel
  methods.
\newblock {\em J. Autom. Reasoning}, 52(2):191--213, 2014.
\newblock \href {http://dx.doi.org/10.1007/s10817-013-9286-5}
  {\path{doi:10.1007/s10817-013-9286-5}}.

\bibitem{BancerekBGKMNP18}
Grzegorz Bancerek, Czeslaw Bylinski, Adam Grabowski, Artur Kornilowicz, Roman
  Matuszewski, Adam Naumowicz, and Karol Pak.
\newblock The role of the {M}izar {M}athematical {L}ibrary for interactive
  proof development in {M}izar.
\newblock {\em J. Autom. Reason.}, 61(1-4):9--32, 2018.
\newblock URL: \url{https://doi.org/10.1007/s10817-017-9440-6}, \href
  {http://dx.doi.org/10.1007/s10817-017-9440-6}
  {\path{doi:10.1007/s10817-017-9440-6}}.

\bibitem{BancerekBGKMNPU15}
Grzegorz Bancerek, Czeslaw Bylinski, Adam Grabowski, Artur Kornilowicz, Roman
  Matuszewski, Adam Naumowicz, Karol Pak, and Josef Urban.
\newblock Mizar: {S}tate-of-the-art and beyond.
\newblock In Manfred Kerber, Jacques Carette, Cezary Kaliszyk, Florian Rabe,
  and Volker Sorge, editors, {\em Intelligent Computer Mathematics -
  International Conference, {CICM} 2015, Washington, DC, USA, July 13-17, 2015,
  Proceedings}, volume 9150 of {\em Lecture Notes in Computer Science}, pages
  261--279. Springer, 2015.
\newblock URL: \url{https://doi.org/10.1007/978-3-319-20615-8\_17}, \href
  {http://dx.doi.org/10.1007/978-3-319-20615-8\_17}
  {\path{doi:10.1007/978-3-319-20615-8\_17}}.

\bibitem{BancerekR02}
Grzegorz Bancerek and Piotr Rudnicki.
\newblock A {Compendium of Continuous Lattices} in {MIZAR}.
\newblock {\em J. Autom. Reasoning}, 29(3-4):189--224, 2002.

\bibitem{BlanchetteGKKU16}
Jasmin~Christian Blanchette, David Greenaway, Cezary Kaliszyk, Daniel
  K{\"{u}}hlwein, and Josef Urban.
\newblock A learning-based fact selector for {Isabelle/HOL}.
\newblock {\em J. Autom. Reasoning}, 57(3):219--244, 2016.
\newblock URL: \url{http://dx.doi.org/10.1007/s10817-016-9362-8}, \href
  {http://dx.doi.org/10.1007/s10817-016-9362-8}
  {\path{doi:10.1007/s10817-016-9362-8}}.

\bibitem{hammers4qed}
Jasmin~Christian Blanchette, Cezary Kaliszyk, Lawrence~C. Paulson, and Josef
  Urban.
\newblock Hammering towards {QED}.
\newblock {\em J. Formalized Reasoning}, 9(1):101--148, 2016.
\newblock URL: \url{http://dx.doi.org/10.6092/issn.1972-5787/4593}, \href
  {http://dx.doi.org/10.6092/issn.1972-5787/4593}
  {\path{doi:10.6092/issn.1972-5787/4593}}.

\bibitem{Chen:2016:XST:2939672.2939785}
Tianqi Chen and Carlos Guestrin.
\newblock {XGBoost}: A scalable tree boosting system.
\newblock In {\em Proceedings of the 22nd ACM SIGKDD International Conference
  on Knowledge Discovery and Data Mining}, KDD '16, pages 785--794, New York,
  NY, USA, 2016. ACM.
\newblock URL: \url{http://doi.acm.org/10.1145/2939672.2939785}, \href
  {http://dx.doi.org/10.1145/2939672.2939785}
  {\path{doi:10.1145/2939672.2939785}}.

\bibitem{Cho14}
Kyunghyun Cho, Bart van Merrienboer, {\c{C}}aglar G{\"{u}}l{\c{c}}ehre, Dzmitry
  Bahdanau, Fethi Bougares, Holger Schwenk, and Yoshua Bengio.
\newblock Learning phrase representations using {RNN} encoder-decoder for
  statistical machine translation.
\newblock In Alessandro Moschitti, Bo~Pang, and Walter Daelemans, editors, {\em
  Proceedings of the 2014 Conference on Empirical Methods in Natural Language
  Processing, {EMNLP} 2014, October 25-29, 2014, Doha, Qatar, {A} meeting of
  SIGDAT, a Special Interest Group of the {ACL}}, pages 1724--1734. {ACL},
  2014.
\newblock URL: \url{http://aclweb.org/anthology/D/D14/D14-1179.pdf}, \href
  {http://dx.doi.org/10.3115/v1/d14-1179} {\path{doi:10.3115/v1/d14-1179}}.

\bibitem{DBLP:conf/tableaux/ChvalovskyJOU21}
Karel Chvalovsk{\'{y}}, Jan Jakubuv, Miroslav Ols{\'{a}}k, and Josef Urban.
\newblock Learning theorem proving components.
\newblock In Anupam Das and Sara Negri, editors, {\em Automated Reasoning with
  Analytic Tableaux and Related Methods - 30th International Conference,
  {TABLEAUX} 2021, Birmingham, UK, September 6-9, 2021, Proceedings}, volume
  12842 of {\em Lecture Notes in Computer Science}, pages 266--278. Springer,
  2021.
\newblock URL: \url{https://doi.org/10.1007/978-3-030-86059-2\_16}, \href
  {http://dx.doi.org/10.1007/978-3-030-86059-2\_16}
  {\path{doi:10.1007/978-3-030-86059-2\_16}}.

\bibitem{ChvalovskyJ0U19}
Karel Chvalovsk{\'{y}}, Jan Jakubuv, Martin Suda, and Josef Urban.
\newblock {ENIGMA-NG:} {E}fficient neural and gradient-boosted inference
  guidance for {E}.
\newblock In Pascal Fontaine, editor, {\em Automated Deduction - {CADE} 27 -
  27th International Conference on Automated Deduction, Natal, Brazil, August
  27-30, 2019, Proceedings}, volume 11716 of {\em Lecture Notes in Computer
  Science}, pages 197--215. Springer, 2019.
\newblock URL: \url{https://doi.org/10.1007/978-3-030-29436-6\_12}, \href
  {http://dx.doi.org/10.1007/978-3-030-29436-6\_12}
  {\path{doi:10.1007/978-3-030-29436-6\_12}}.

\bibitem{Evans18}
Richard Evans, David Saxton, David Amos, Pushmeet Kohli, and Edward
  Grefenstette.
\newblock Can neural networks understand logical entailment?
\newblock In {\em International Conference on Learning Representations}, 2018.
\newblock URL: \url{https://openreview.net/forum?id=SkZxCk-0Z}.

\bibitem{DBLP:conf/lpar/Gauthier20}
Thibault Gauthier.
\newblock Deep reinforcement learning for synthesizing functions in
  higher-order logic.
\newblock In {\em {LPAR}}, volume~73 of {\em EPiC Series in Computing}, pages
  230--248. EasyChair, 2020.

\bibitem{GauthierKU17}
Thibault Gauthier, Cezary Kaliszyk, and Josef Urban.
\newblock {TacticToe}: Learning to reason with {HOL4} tactics.
\newblock In Thomas Eiter and David Sands, editors, {\em LPAR-21, 21st
  International Conference on Logic for Programming, Artificial Intelligence
  and Reasoning, Maun, Botswana, May 7-12, 2017}, volume~46 of {\em EPiC},
  pages 125--143. EasyChair, 2017.
\newblock URL: \url{https://doi.org/10.29007/ntlb}, \href
  {http://dx.doi.org/10.29007/ntlb} {\path{doi:10.29007/ntlb}}.

\bibitem{GauthierKUKN21}
Thibault Gauthier, Cezary Kaliszyk, Josef Urban, Ramana Kumar, and Michael
  Norrish.
\newblock Tactictoe: Learning to prove with tactics.
\newblock {\em J. Autom. Reason.}, 65(2):257--286, 2021.
\newblock URL: \url{https://doi.org/10.1007/s10817-020-09580-x}, \href
  {http://dx.doi.org/10.1007/s10817-020-09580-x}
  {\path{doi:10.1007/s10817-020-09580-x}}.

\bibitem{abs-2301-11479}
Thibault Gauthier, Miroslav Ols{\'{a}}k, and Josef Urban.
\newblock Alien coding.
\newblock {\em CoRR}, abs/2301.11479, 2023.

\bibitem{abs-2202-11908}
Thibault Gauthier and Josef Urban.
\newblock Learning program synthesis for integer sequences from scratch.
\newblock {\em CoRR}, abs/2202.11908, 2022.

\bibitem{10.1007/978-3-030-29026-9_21}
Zarathustra Goertzel, Jan Jakub{\r{u}}v, and Josef Urban.
\newblock {ENIGMAWatch: ProofWatch} meets {ENIGMA}.
\newblock In Serenella Cerrito and Andrei Popescu, editors, {\em Automated
  Reasoning with Analytic Tableaux and Related Methods}, pages 374--388, Cham,
  2019. Springer International Publishing.

\bibitem{Goertzel20}
Zarathustra~Amadeus Goertzel.
\newblock Make {E} smart again (short paper).
\newblock In {\em {IJCAR} {(2)}}, volume 12167 of {\em Lecture Notes in
  Computer Science}, pages 408--415. Springer, 2020.

\bibitem{GoertzelCJOU21}
Zarathustra~Amadeus Goertzel, Karel Chvalovsk{\'{y}}, Jan Jakubuv, Miroslav
  Ols{\'{a}}k, and Josef Urban.
\newblock Fast and slow {E}nigmas and parental guidance.
\newblock In {\em FroCoS}, volume 12941 of {\em Lecture Notes in Computer
  Science}, pages 173--191. Springer, 2021.

\bibitem{GoertzelJKOPU22}
Zarathustra~Amadeus Goertzel, Jan Jakubuv, Cezary Kaliszyk, Miroslav
  Ols{\'{a}}k, Jelle Piepenbrock, and Josef Urban.
\newblock The {I}sabelle {ENIGMA}.
\newblock In {\em {ITP}}, volume 237 of {\em LIPIcs}, pages 16:1--16:21.
  Schloss Dagstuhl - Leibniz-Zentrum f{\"{u}}r Informatik, 2022.

\bibitem{mizar-in-a-nutshell}
Adam Grabowski, Artur Korni{\l}owicz, and Adam Naumowicz.
\newblock {M}izar in a nutshell.
\newblock {\em J. Formalized Reasoning}, 3(2):153--245, 2010.

\bibitem{ParamILS-JAIR}
Frank Hutter, Holger~H. Hoos, Kevin Leyton-Brown, and Thomas St\"{u}tzle.
\newblock {ParamILS:} an automatic algorithm configuration framework.
\newblock {\em J. Artificial Intelligence Research}, 36:267--306, October 2009.

\bibitem{JakubuvU18a}
Jan Jakub\r{u}v and Josef Urban.
\newblock Hierarchical invention of theorem proving strategies.
\newblock {\em {AI} Commun.}, 31(3):237--250, 2018.
\newblock URL: \url{https://doi.org/10.3233/AIC-180761}, \href
  {http://dx.doi.org/10.3233/AIC-180761} {\path{doi:10.3233/AIC-180761}}.

\bibitem{JakubuvCOP0U20}
Jan Jakubuv, Karel Chvalovsk{\'{y}}, Miroslav Ols{\'{a}}k, Bartosz Piotrowski,
  Martin Suda, and Josef Urban.
\newblock {ENIGMA} anonymous: Symbol-independent inference guiding machine
  (system description).
\newblock In {\em {IJCAR} {(2)}}, volume 12167 of {\em Lecture Notes in
  Computer Science}, pages 448--463. Springer, 2020.

\bibitem{JakubuvSU17}
Jan Jakubuv, Martin Suda, and Josef Urban.
\newblock Automated invention of strategies and term orderings for vampire.
\newblock In {\em {GCAI}}, volume~50 of {\em EPiC Series in Computing}, pages
  121--133. EasyChair, 2017.

\bibitem{JakubuvU17}
Jan Jakubuv and Josef Urban.
\newblock {BliStrTune}: hierarchical invention of theorem proving strategies.
\newblock In Yves Bertot and Viktor Vafeiadis, editors, {\em Proceedings of the
  6th {ACM} {SIGPLAN} Conference on Certified Programs and Proofs, {CPP} 2017,
  Paris, France, January 16-17, 2017}, pages 43--52. {ACM}, 2017.
\newblock URL: \url{http://doi.acm.org/10.1145/3018610.3018619}, \href
  {http://dx.doi.org/10.1145/3018610.3018619}
  {\path{doi:10.1145/3018610.3018619}}.

\bibitem{JakubuvU18}
Jan Jakubuv and Josef Urban.
\newblock Enhancing {ENIGMA} given clause guidance.
\newblock In Florian Rabe, William~M. Farmer, Grant~O. Passmore, and Abdou
  Youssef, editors, {\em Intelligent Computer Mathematics - 11th International
  Conference, {CICM} 2018, Hagenberg, Austria, August 13-17, 2018,
  Proceedings}, volume 11006 of {\em Lecture Notes in Computer Science}, pages
  118--124. Springer, 2018.
\newblock URL: \url{https://doi.org/10.1007/978-3-319-96812-4\_11}, \href
  {http://dx.doi.org/10.1007/978-3-319-96812-4\_11}
  {\path{doi:10.1007/978-3-319-96812-4\_11}}.

\bibitem{JakubuvU19}
Jan Jakubuv and Josef Urban.
\newblock Hammering {Mizar} by learning clause guidance.
\newblock In John Harrison, John O'Leary, and Andrew Tolmach, editors, {\em
  10th International Conference on Interactive Theorem Proving, {ITP} 2019,
  September 9-12, 2019, Portland, OR, {USA}}, volume 141 of {\em LIPIcs}, pages
  34:1--34:8. Schloss Dagstuhl - Leibniz-Zentrum f{\"{u}}r Informatik, 2019.
\newblock URL: \url{https://doi.org/10.4230/LIPIcs.ITP.2019.34}, \href
  {http://dx.doi.org/10.4230/LIPIcs.ITP.2019.34}
  {\path{doi:10.4230/LIPIcs.ITP.2019.34}}.

\bibitem{EasyChair:74}
Cezary Kaliszyk and Josef Urban.
\newblock Stronger automation for {F}lyspeck by feature weighting and strategy
  evolution.
\newblock In Jasmin~Christian Blanchette and Josef Urban, editors, {\em PxTP
  2013}, volume~14 of {\em EPiC Series}, pages 87--95. EasyChair, 2013.

\bibitem{holyhammer}
Cezary Kaliszyk and Josef Urban.
\newblock Learning-assisted automated reasoning with {F}lyspeck.
\newblock {\em J. Autom. Reasoning}, 53(2):173--213, 2014.
\newblock URL: \url{http://dx.doi.org/10.1007/s10817-014-9303-3}, \href
  {http://dx.doi.org/10.1007/s10817-014-9303-3}
  {\path{doi:10.1007/s10817-014-9303-3}}.

\bibitem{KaliszykU13b}
Cezary Kaliszyk and Josef Urban.
\newblock {MizAR 40 for Mizar 40}.
\newblock {\em J. Autom. Reasoning}, 55(3):245--256, 2015.
\newblock URL: \url{http://dx.doi.org/10.1007/s10817-015-9330-8}, \href
  {http://dx.doi.org/10.1007/s10817-015-9330-8}
  {\path{doi:10.1007/s10817-015-9330-8}}.

\bibitem{DBLP:conf/ijcai/KaliszykUV15}
Cezary Kaliszyk, Josef Urban, and Jir{\'{\i}} Vyskocil.
\newblock Efficient semantic features for automated reasoning over large
  theories.
\newblock In {\em {IJCAI}}, pages 3084--3090. {AAAI} Press, 2015.

\bibitem{DBLP:conf/itp/KaliszykUV17}
Cezary Kaliszyk, Josef Urban, and Jir{\'{\i}} Vyskocil.
\newblock Automating formalization by statistical and semantic parsing of
  mathematics.
\newblock In {\em {ITP}}, volume 10499 of {\em Lecture Notes in Computer
  Science}, pages 12--27. Springer, 2017.

\bibitem{LightGBM}
Guolin Ke, Qi~Meng, Thomas Finley, Taifeng Wang, Wei Chen, Weidong Ma, Qiwei
  Ye, and Tie{-}Yan Liu.
\newblock Lightgbm: {A} highly efficient gradient boosting decision tree.
\newblock In {\em {NIPS}}, pages 3146--3154, 2017.

\bibitem{SCM}
Artur Korniłowicz and Christoph Schwarzweller.
\newblock Computers and algorithms in {M}izar.
\newblock {\em Mechanized Mathematics and Its Applications}, 4(1):43--50, 2005.

\bibitem{Vampire}
Laura Kov{\'a}cs and Andrei Voronkov.
\newblock First-order theorem proving and {V}ampire.
\newblock In Natasha Sharygina and Helmut Veith, editors, {\em CAV}, volume
  8044 of {\em LNCS}, pages 1--35. Springer, 2013.

\bibitem{Lample20}
Guillaume Lample and Fran{\c{c}}ois Charton.
\newblock Deep learning for symbolic mathematics.
\newblock In {\em 8th International Conference on Learning Representations,
  {ICLR} 2020, Addis Ababa, Ethiopia, April 26-30, 2020}. OpenReview.net, 2020.
\newblock URL: \url{https://openreview.net/forum?id=S1eZYeHFDS}.

\bibitem{OlsakKU20}
Miroslav Ols{\'{a}}k, Cezary Kaliszyk, and Josef Urban.
\newblock Property invariant embedding for automated reasoning.
\newblock In Giuseppe~De Giacomo, Alejandro Catal{\'{a}}, Bistra Dilkina,
  Michela Milano, Sen{\'{e}}n Barro, Alberto Bugar{\'{\i}}n, and
  J{\'{e}}r{\^{o}}me Lang, editors, {\em {ECAI} 2020 - 24th European Conference
  on Artificial Intelligence}, volume 325 of {\em Frontiers in Artificial
  Intelligence and Applications}, pages 1395--1402. {IOS} Press, 2020.
\newblock URL: \url{https://doi.org/10.3233/FAIA200244}, \href
  {http://dx.doi.org/10.3233/FAIA200244} {\path{doi:10.3233/FAIA200244}}.

\bibitem{OpitzM99}
David~W. Opitz and Richard Maclin.
\newblock Popular ensemble methods: An empirical study.
\newblock {\em J. Artif. Intell. Res.}, 11:169--198, 1999.
\newblock URL: \url{https://doi.org/10.1613/jair.614}, \href
  {http://dx.doi.org/10.1613/jair.614} {\path{doi:10.1613/jair.614}}.

\bibitem{abs-2210-03590}
Jelle Piepenbrock, Josef Urban, Konstantin Korovin, Miroslav Ols{\'{a}}k, Tom
  Heskes, and Mikolas Janota.
\newblock Machine learning meets the herbrand universe.
\newblock {\em CoRR}, abs/2210.03590, 2022.

\bibitem{PiotrowskiU18}
Bartosz Piotrowski and Josef Urban.
\newblock {ATPboost}: Learning premise selection in binary setting with {ATP}
  feedback.
\newblock In Didier Galmiche, Stephan Schulz, and Roberto Sebastiani, editors,
  {\em Automated Reasoning - 9th International Joint Conference, {IJCAR} 2018,
  Held as Part of the Federated Logic Conference, FloC 2018, Oxford, UK, July
  14-17, 2018, Proceedings}, volume 10900 of {\em Lecture Notes in Computer
  Science}, pages 566--574. Springer, 2018.
\newblock URL: \url{https://doi.org/10.1007/978-3-319-94205-6\_37}, \href
  {http://dx.doi.org/10.1007/978-3-319-94205-6\_37}
  {\path{doi:10.1007/978-3-319-94205-6\_37}}.

\bibitem{PiotrowskiU20}
Bartosz Piotrowski and Josef Urban.
\newblock Stateful premise selection by recurrent neural networks.
\newblock In {\em {LPAR}}, volume~73 of {\em EPiC Series in Computing}, pages
  409--422. EasyChair, 2020.

\bibitem{abs-1911-04873}
Bartosz Piotrowski, Josef Urban, Chad~E. Brown, and Cezary Kaliszyk.
\newblock Can neural networks learn symbolic rewriting?
\newblock {\em CoRR}, abs/1911.04873, 2019.

\bibitem{DBLP:books/el/RobinsonV01}
John~Alan Robinson and Andrei Voronkov, editors.
\newblock {\em Handbook of Automated Reasoning (in 2 volumes)}.
\newblock Elsevier and {MIT} Press, 2001.

\bibitem{Sch02-AICOMM}
Stephan Schulz.
\newblock {E -- A Brainiac Theorem Prover}.
\newblock {\em AI Commun.}, 15(2-3):111--126, 2002.

\bibitem{DBLP:conf/birthday/Schulz13}
Stephan Schulz.
\newblock Simple and efficient clause subsumption with feature vector indexing.
\newblock In {\em Automated Reasoning and Mathematics}, volume 7788 of {\em
  Lecture Notes in Computer Science}, pages 45--67. Springer, 2013.

\bibitem{Schulz13}
Stephan Schulz.
\newblock System description: {E} 1.8.
\newblock In Kenneth~L. McMillan, Aart Middeldorp, and Andrei Voronkov,
  editors, {\em LPAR}, volume 8312 of {\em LNCS}, pages 735--743. Springer,
  2013.
\newblock URL: \url{http://dx.doi.org/10.1007/978-3-642-45221-5_49}, \href
  {http://dx.doi.org/10.1007/978-3-642-45221-5_49}
  {\path{doi:10.1007/978-3-642-45221-5_49}}.

\bibitem{Schulz:LPAR-2013}
Stephan Schulz.
\newblock {System Description: E~1.8}.
\newblock In Ken McMillan, Aart Middeldorp, and Andrei Voronkov, editors, {\em
  Proc.\ of the 19th LPAR, Stellenbosch}, volume 8312 of {\em LNCS}, pages
  735--743. Springer, 2013.

\bibitem{SCV:CADE-2019}
Stephan Schulz, Simon Cruanes, and Petar Vukmirovi{\'c}.
\newblock Faster, higher, stronger: {E} 2.3.
\newblock In Pascal Fontaine, editor, {\em Proc.\ of the 27th CADE, Natal,
  Brasil}, number 11716 in LNAI, pages 495--507. Springer, 2019.

\bibitem{SS:APPA-2015}
Stephan Schulz and Geoff Sutcliffe.
\newblock Proof generation for saturating first-order theorem provers.
\newblock In David Delahaye and Bruno Woltzenlogel~Paleo, editors, {\em {All
  about Proofs, Proofs for All}}, volume~55 of {\em Mathematical Logic and
  Foundations}, pages 45--61. College Publications, London, UK, January 2015.

\bibitem{DBLP:conf/cade/000121a}
Martin Suda.
\newblock Improving {ENIGMA}-style clause selection while learning from
  history.
\newblock In Andr{\'{e}} Platzer and Geoff Sutcliffe, editors, {\em Automated
  Deduction - {CADE} 28 - 28th International Conference on Automated Deduction,
  Virtual Event, July 12-15, 2021, Proceedings}, volume 12699 of {\em Lecture
  Notes in Computer Science}, pages 543--561. Springer, 2021.
\newblock URL: \url{https://doi.org/10.1007/978-3-030-79876-5\_31}, \href
  {http://dx.doi.org/10.1007/978-3-030-79876-5\_31}
  {\path{doi:10.1007/978-3-030-79876-5\_31}}.

\bibitem{DBLP:conf/frocos/Suda21}
Martin Suda.
\newblock Vampire with a brain is a good {ITP} hammer.
\newblock In Boris Konev and Giles Reger, editors, {\em Frontiers of Combining
  Systems - 13th International Symposium, FroCoS 2021, Birmingham, UK,
  September 8-10, 2021, Proceedings}, volume 12941 of {\em Lecture Notes in
  Computer Science}, pages 192--209. Springer, 2021.
\newblock URL: \url{https://doi.org/10.1007/978-3-030-86205-3\_11}, \href
  {http://dx.doi.org/10.1007/978-3-030-86205-3\_11}
  {\path{doi:10.1007/978-3-030-86205-3\_11}}.

\bibitem{Tammet98}
Tanel Tammet.
\newblock Towards efficient subsumption.
\newblock In {\em {CADE}}, volume 1421 of {\em Lecture Notes in Computer
  Science}, pages 427--441. Springer, 1998.

\bibitem{Urb03}
J.~Urban.
\newblock {Translating {Mizar} for First Order Theorem Provers}.
\newblock In A.~Asperti, B.~Buchberger, and J.H. Davenport, editors, {\em
  {Proceedings of the 2nd International Conference on Mathematical Knowledge
  Management}}, number 2594 in LNCS, pages 203--215. Springer, 2003.

\bibitem{Urb04-MPTP0}
Josef Urban.
\newblock {MPTP -- Motivation, Implementation, First Experiments}.
\newblock {\em J. Autom. Reasoning}, 33(3-4):319--339, 2004.
\newblock \href {http://dx.doi.org/10.1007/s10817-004-6245-1}
  {\path{doi:10.1007/s10817-004-6245-1}}.

\bibitem{Urban06}
Josef Urban.
\newblock {MPTP} 0.2: Design, implementation, and initial experiments.
\newblock {\em J. Autom. Reasoning}, 37(1-2):21--43, 2006.

\bibitem{blistr}
Josef Urban.
\newblock {BliStr: The Blind Strategymaker}.
\newblock In Georg Gottlob, Geoff Sutcliffe, and Andrei Voronkov, editors, {\em
  Global Conference on Artificial Intelligence, {GCAI} 2015, Tbilisi, Georgia,
  October 16-19, 2015}, volume~36 of {\em EPiC Series in Computing}, pages
  312--319. EasyChair, 2015.
\newblock URL:
  \url{http://www.easychair.org/publications/paper/BliStr_The_Blind_Strategymaker}.

\bibitem{UrbanJ20}
Josef Urban and Jan Jakubuv.
\newblock First neural conjecturing datasets and experiments.
\newblock In Christoph Benzm{\"{u}}ller and Bruce~R. Miller, editors, {\em
  Intelligent Computer Mathematics - 13th International Conference, {CICM}
  2020, Bertinoro, Italy, July 26-31, 2020, Proceedings}, volume 12236 of {\em
  Lecture Notes in Computer Science}, pages 315--323. Springer, 2020.
\newblock URL: \url{https://doi.org/10.1007/978-3-030-53518-6\_24}, \href
  {http://dx.doi.org/10.1007/978-3-030-53518-6\_24}
  {\path{doi:10.1007/978-3-030-53518-6\_24}}.

\bibitem{US+08}
Josef Urban, Geoff Sutcliffe, Petr Pudl{\'a}k, and Ji\v{r}\'{\i} Vysko\v{c}il.
\newblock {MaLARea SG1 -- Machine Learner for Automated Reasoning with Semantic
  Guidance}.
\newblock In {\em IJCAR}, pages 441--456, 2008.

\bibitem{DBLP:journals/corr/VaswaniSPUJGKP17}
Ashish Vaswani, Noam Shazeer, Niki Parmar, Jakob Uszkoreit, Llion Jones,
  Aidan~N. Gomez, Lukasz Kaiser, and Illia Polosukhin.
\newblock Attention is all you need.
\newblock {\em CoRR}, abs/1706.03762, 2017.
\newblock URL: \url{http://arxiv.org/abs/1706.03762}, \href
  {http://arxiv.org/abs/1706.03762} {\path{arXiv:1706.03762}}.

\bibitem{Wang18}
Qingxiang Wang, Cezary Kaliszyk, and Josef Urban.
\newblock First experiments with neural translation of informal to formal
  mathematics.
\newblock In Florian Rabe, William~M. Farmer, Grant~O. Passmore, and Abdou
  Youssef, editors, {\em 11th International Conference on Intelligent Computer
  Mathematics (CICM 2018)}, volume 11006 of {\em LNCS}, pages 255--270.
  Springer, 2018.
\newblock URL: \url{https://doi.org/10.1007/978-3-319-96812-4\_22}, \href
  {http://dx.doi.org/10.1007/978-3-319-96812-4\_22}
  {\path{doi:10.1007/978-3-319-96812-4\_22}}.

\end{thebibliography}

\appendix
\section{Further Proof Details and Discussion}
\subsection{Further proof details of NEWTON:72}
\begin{sidewaysfigure}[ht]
\begin{center}
  \includegraphics[width=1\textwidth]{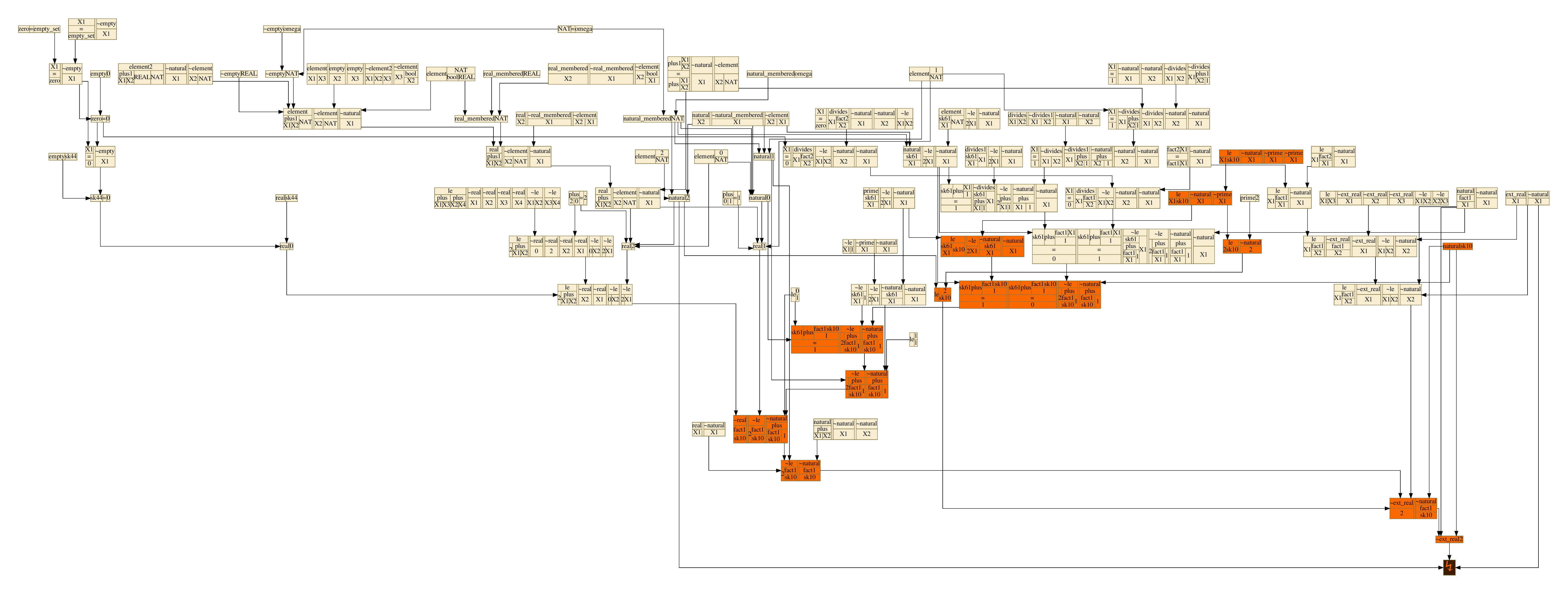}
\end{center}
\caption{ENIGMA's proof of NEWTON:72.}
\label{fig:t72newton}
\end{sidewaysfigure}

We show the full proof of \th{NEWTON:72} -- \emph{For every natural number there is a larger prime} --
without the clausification steps in Figure~\ref{fig:t72newton}.
The verbatim ENIGMA proof is available online.\footnoteA{\url{https://bit.ly/3Z2iXo3}}
The original conjecture is as follows. Note that stating twice the primality is an artifact of the Mizar encoding.
\begin{small}
  \begin{verbatim}
fof(t72_newton,conjecture,(
    ! [X1] :
      ( natural(X1)
     => ? [X2] :
          ( natural(X2)
          & prime(X2)
          & prime(X2)
          & ~ le(X2,X1) ) ) ),
    file(t72_newton,t72_newton)).
\end{verbatim}
\end{small}
Then the conjecture is negated. It is not the case that for every (natural number) X1 there is a prime
(natural number) X2 with $X2 > X1$.
\begin{small}
  \begin{verbatim}
fof(c_0_54,negated_conjecture,(
    ~ ( ! [X1] :
          ( natural(X1)
         => ? [X2] :
              ( natural(X2)
              & prime(X2)
              & prime(X2)
              & ~ le(X2,X1) ) ) ) ),
    inference(assume_negation,[status(cth)],[t72_newton])).
\end{verbatim}
\end{small}
After that the X1 (the number a bigger prime exists for) gets
skolemized into esk1\_0. Since this is now negated, instead of ``there
is a X6'' we get a ``for all X6'' with  X6 is not prime, or X6 is $\le$
that number.
\begin{small}
  \begin{verbatim}
fof(c_0_64,negated_conjecture,(
    ! [X6] :
      ( natural(esk1_0)
      & ( ~ natural(X6)
        | ~ prime(X6)
        | ~ prime(X6)
        | le(X6,esk1_0) ) ) ),
    inference(shift_quantors,[status(thm)],[inference(skolemize,[status(esa)],...)]).
\end{verbatim}
\end{small}
In the following inference, the prover synthesizes the term $x1!+1$
``plus(fact1(X1),n1)'' (called \emph{witness} from now on) for the first time.
\begin{small}
  \begin{verbatim}
cnf(c_0_96,plain,
    ( esk60_1(plus(fact1(X1),n1)) = n0
    | esk60_1(plus(fact1(X1),n1)) = n1
    | ~ le(esk60_1(plus(fact1(X1),n1)),X1)
    | ~ le(n2,plus(fact1(X1),n1))
    | ~ natural(plus(fact1(X1),n1))
    | ~ natural(X1) ),
    inference(csr,[status(thm)],[inference(csr,[status(thm)],[inference(...)])]).
\end{verbatim}
\end{small}
This then goes into the conjecture here:
\begin{small}
  \begin{verbatim}
cnf(c_0_106,negated_conjecture,
    ( esk60_1(plus(fact1(esk1_0),n1)) = n1
    | esk60_1(plus(fact1(esk1_0),n1)) = n0
    | ~ le(n2,plus(fact1(esk1_0),n1))
    | ~ natural(plus(fact1(esk1_0),n1)) ),
    inference(csr,[status(thm)],[inference(cn,[status(thm)]...)]).
\end{verbatim}
\end{small}
which essentially says:
\begin{small}
  \begin{verbatim}
    witness=1
 or witness=0
 or witness>2 ("not witness<=2")
 or witness is not a natural number
\end{verbatim}
\end{small}
Once this is established, the four cases are handled in relatively straightforward way in the proof.

\subsection{NEWTON:72 as an example of ML-guided deductive synthesis}
A simplified core behind the above-explained synthesis of the nontrivial witness X1!+1 in the proof 
is the combination of the following two Mizar theorems \th{NEWTON:39} and  \th{NEWTON:41}:
\begin{small}
  \begin{verbatim}
theorem Th39: :: NEWTON:39
for m, n being Nat st m <> 1 & m divides n holds not m divides n + 1

theorem Th41: :: NEWTON:41
for j, l being Nat st j <= l & j <> 0 holds j divides l !
\end{verbatim}
\end{small}
  When resolved, the unification leads to essentially substituting
  \texttt{l!} for n in \th{NEWTON:39}, and thus synthesizing the
  instance \texttt{l!+1} which is the required witness (j divides l!, hence it indeed cannot divide l!+1).

This 
  demonstrates how (A) (learning-guided) resolution-based synthesis
  differs from (B) pure learning-guided synthesis or enumeration.
  In
  (A), the learner recommends the clauses that should be resolved
  (parental guidance) or the single (given) clause that is relevant
  wrt. a set of already selected clauses (standard GBDT/GNN
  guidance). This recommendation may or may not fully understand what
  will be the result of the resolution. It is quite likely just a
  ``hunch'' that combining the two facts may be interesting enough,
  and the actual deductively synthesized witness comes as a surprise that
  is here ``computed'' by the logical calculus.

  Whereas in (B), we require
  the learner (e.g. a language model~\cite{UrbanJ20}, GNN2RNN~\cite{abs-2210-03590}, or an
  MCTS-guided synthesis framework~\cite{abs-2202-11908}) to ``come up with the term
  on its own, based on looking at the situation''. This is plausible in situations where the
  mathematician has already seen and recognizes the pattern/trick, or
  analogizes, transferring the terminology as in GNN2RNN. The
  better and larger the pattern recognition and analogizing database
  and capability, the surer the mathematician will be in directly
  coming up with the witness.

  Approach (A) thus seems more exploratory and applicable to new
  problem-solving situations, reinforcing weaker hunches by deduction/computation
  leading to possibly novel and surprising discoveries and values. While approach (B)
  seems useful in known situations, relying more on direct and
  sufficiently confident recall and memorization of the likely values.
  Approach (A) is more a hunch about the method or direction that
  should be used. E.g., we may feel we should somehow combine results X, Y,
  and Z, compute a particular derivative/integral, solve a set of
  equations/constraints, look for an inductive hypothesis,
  etc., without knowing the result. While approach (B) is more a
  direct hunch/recall of the solution rather than of the process
  leading to it.

  There is likely again a feedback loop between Type-A and Type-B
  ML-guided problem-solving approaches. Type-A is more indirect and works by pointing to and
  invoking further uncertain procedures and searching with them. If
  reasonably successful in a particular kind of situation, the
  confidences rise and may lead all the way to more direct Type-B
  guesses. Which when successful may shorten parts of the Type-A
  searches, making those more successful too, etc.

\subsection{Further Sample Proof Graphs}

\begin{sidewaysfigure}[ht]
\begin{center}
  \includegraphics[width=0.8\textwidth]{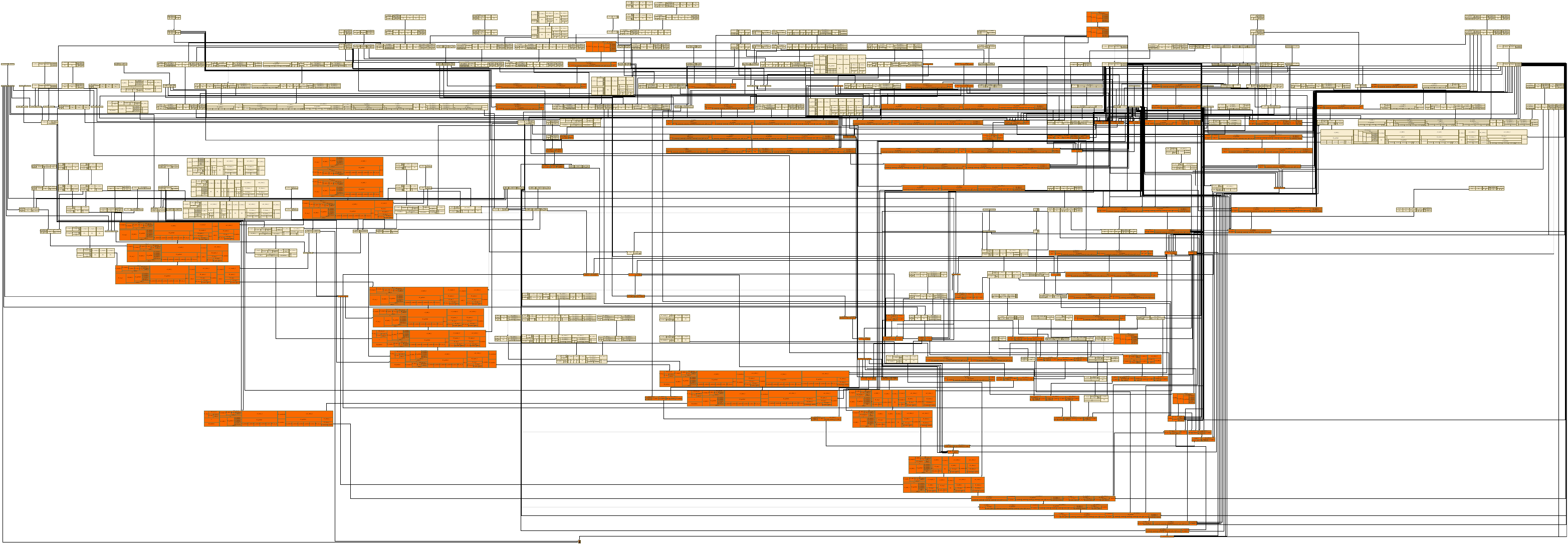}
\end{center}
\caption{ENIGMA's proof of \th{FDIFF_8:14}.}
\label{fig:t14-fdiff-8}
\end{sidewaysfigure}

\begin{sidewaysfigure}[ht]
\begin{center}
  \includegraphics[width=0.7\textwidth]{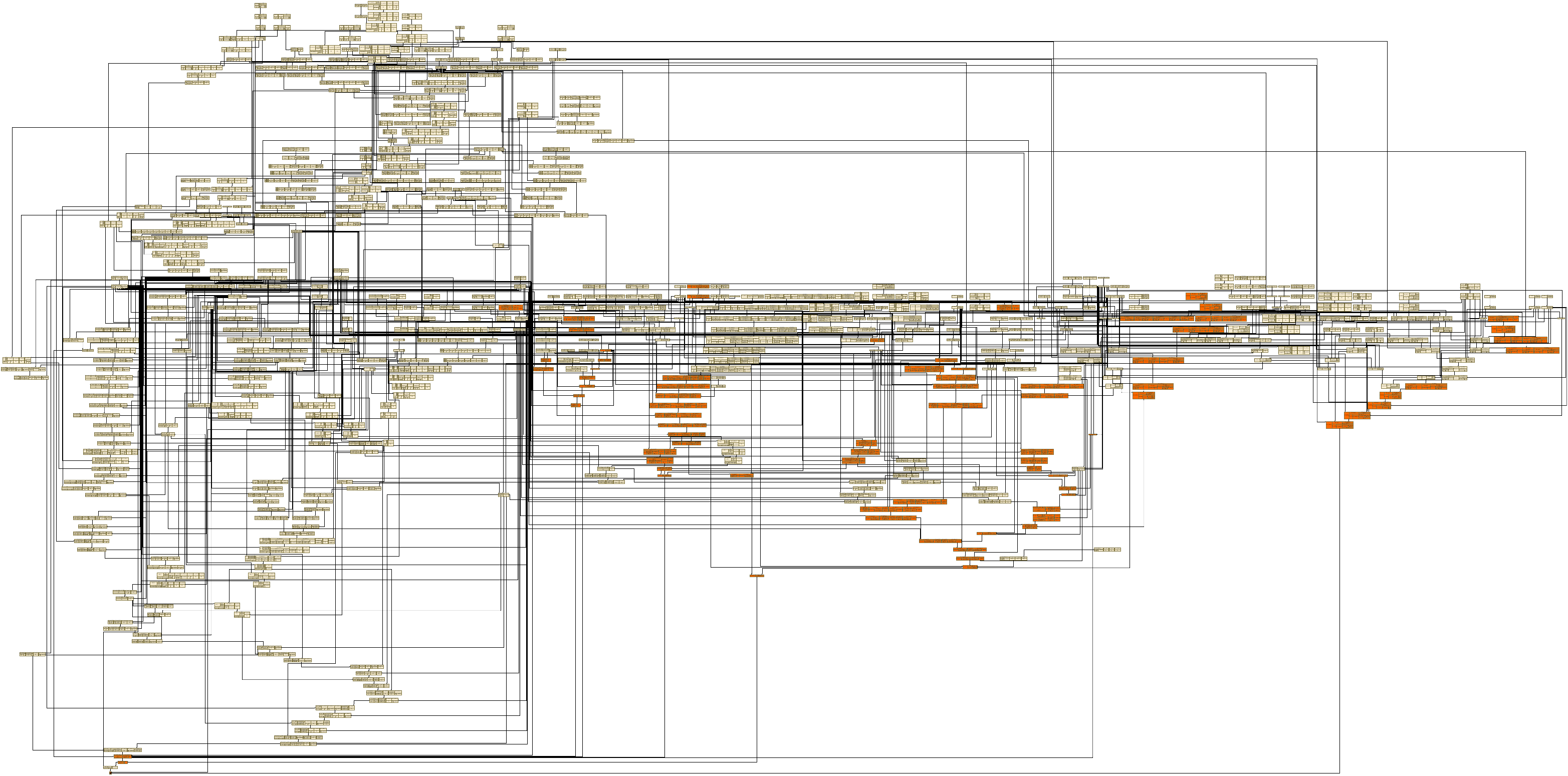}
\end{center}
\caption{ENIGMA's proof of \th{SINCOS10:86}.}
\label{fig:t86-sincos10}
\end{sidewaysfigure}

\begin{sidewaysfigure}[ht]
\begin{center}
  \includegraphics[width=0.8\textwidth]{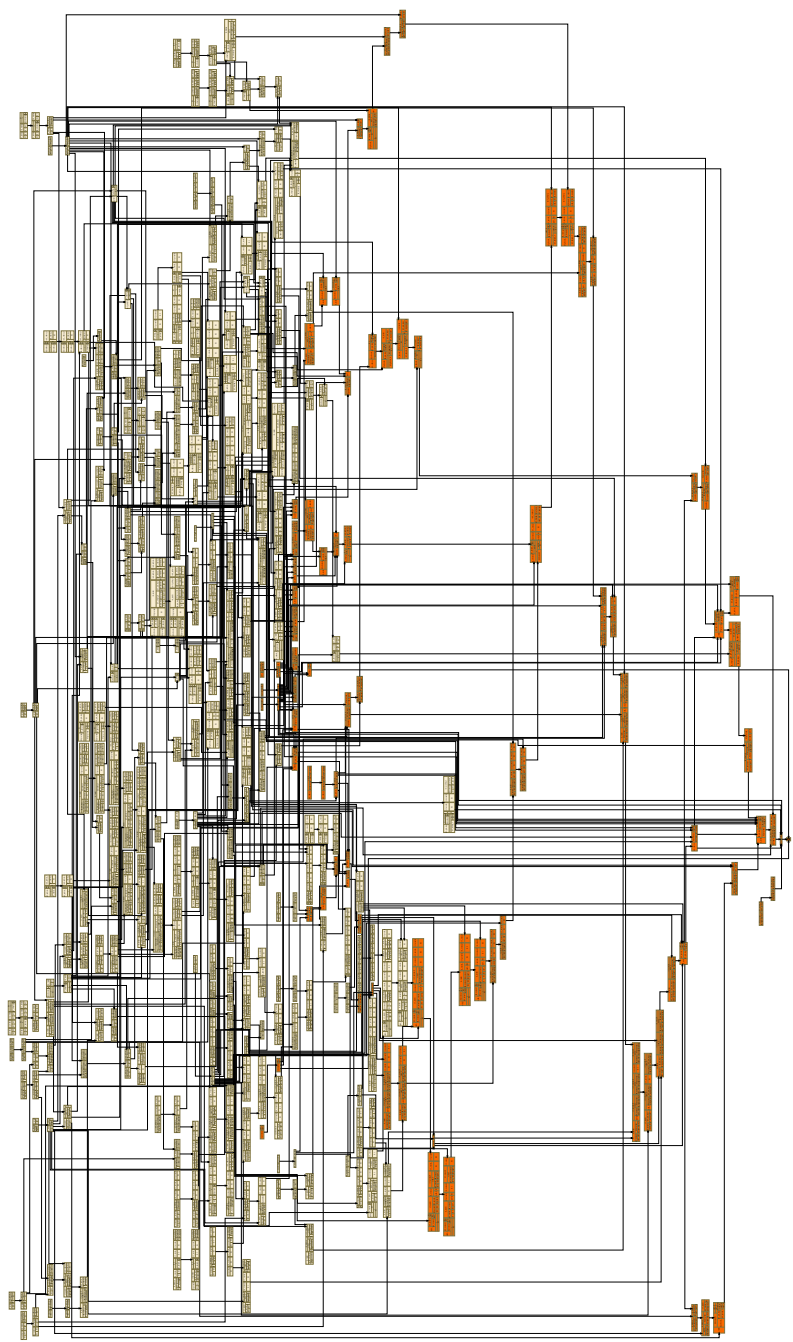}
\end{center}
\caption{ENIGMA's proof of \th{BORSUK_5:31}.}
\label{fig:t31-borsuk-5}
\end{sidewaysfigure}

To give the readers a sample of the complexity of the more advanced
ENIGMA proofs described in Section~\ref{Proofs}, we include the
graphical representation of three of them here (Figure~\ref{fig:t14-fdiff-8}, \ref{fig:t86-sincos10}, \ref{fig:t31-borsuk-5}). We again omit the
clausification part from them. Since these are large graphs with
hundreds of nodes (which may further consist of quite complex
clauses), we refer interested readers also to our web page where the
graphs can be viewed interactively as SVG images.\footnote{\url{http://grid01.ciirc.cvut.cz/~mptp/enigma_prf_graph/t72_newton_nice.proof1.svg}}\footnote{\url{http://grid01.ciirc.cvut.cz/~mptp/enigma_prf_graph/t31_borsuk_5.proof1.svg}}\footnote{\url{http://grid01.ciirc.cvut.cz/~mptp/enigma_prf_graph/t14_fdiff_8.proof1.svg}}\footnote{\url{http://grid01.ciirc.cvut.cz/~mptp/enigma_prf_graph/t86_sincos10.proof1.svg}}\footnote{\url{http://grid01.ciirc.cvut.cz/~mptp/enigma_prf_graph/t22_ideal_1.proof1.svg}}

\end{document}